\documentclass{article}

\usepackage{microtype}
\usepackage{graphicx}
\usepackage{hyperref}       
\usepackage{booktabs} 
\usepackage{multicol}
\usepackage{multirow, varwidth}
\usepackage{amssymb}
\usepackage[linesnumbered,ruled,noend]{algorithm2e}
\usepackage{amsmath}
\usepackage{amsthm}
\usepackage{algpseudocode}
\usepackage[dvipsnames]{xcolor}
\usepackage{colortbl}
\usepackage{natbib}
\usepackage{subcaption}
\usepackage{wrapfig}
\usepackage{xspace} 
\usepackage{enumitem}
\usepackage{tikz}
\usetikzlibrary{tikzmark,calc}
\usepackage{soul}

\definecolor{myblue}{HTML}{5b5bcd}
\definecolor{myorange}{HTML}{fb9902}

\usepackage{hyperref}

\usepackage[accepted]{icml2022}

\newcommand{\name}{{\texttt{AdaDPS}}\xspace}

\newtheorem{definition}{Definition}

\newtheorem{assumption}{Assumption}

\newtheorem{theorem}{Theorem}
\newtheorem*{theorem*}{Theorem}

\newtheorem*{lemma*}{Lemma}

\newtheorem*{corollary*}{Corollary}

\icmltitlerunning{Private Adaptive Optimization with Side information}

\begin{document}

\twocolumn[
\icmltitle{Private Adaptive Optimization with Side information}

\icmlsetsymbol{equal}{*}

\begin{icmlauthorlist}
\icmlauthor{Tian Li}{cmu}
\icmlauthor{Manzil Zaheer}{deepmind}
\icmlauthor{Sashank J. Reddi}{google}
\icmlauthor{Virginia Smith}{cmu}
\end{icmlauthorlist}

\icmlaffiliation{cmu}{Carnegie Mellon University}
\icmlaffiliation{google}{Google Research}
\icmlaffiliation{deepmind}{Google DeepMind}

\icmlcorrespondingauthor{Tian Li}{tianli@cmu.edu}


\vskip 0.3in
]

\printAffiliationsAndNotice{}  

\begin{abstract}
Adaptive optimization methods have become the default solvers for many machine learning tasks. Unfortunately, the benefits of adaptivity may degrade when training with differential privacy, as the noise added to ensure privacy reduces the effectiveness of the adaptive preconditioner. To this end, we propose \name, a general framework that uses \emph{non-sensitive side information} to precondition the gradients, allowing the effective use of adaptive methods in private settings. We formally show \name reduces the amount of noise needed to achieve similar privacy guarantees, thereby improving optimization performance. Empirically, we leverage simple and readily available side information to explore the performance of \name in practice, comparing to strong baselines in both centralized and federated settings. Our results show that \name improves accuracy by 7.7\% (absolute) on average---yielding state-of-the-art privacy-utility trade-offs on  large-scale text and image benchmarks. 
\end{abstract}

\section{Introduction}
\label{sec:intro}

Privacy-sensitive applications in areas such as healthcare and cross-device federated learning have fueled a demand for optimization methods that ensure \textit{differential privacy (DP)}~\cite{dwork2006calibrating,chaudhuri2011differentially,abadi2016deep,mcmahan2017learning}.
These methods typically perturb gradients with random noise at each iteration in order to mask the influence of individual examples on the trained model. As the amount of privacy is directly related to the number of training iterations, private applications stand to benefit from optimizers that improve convergence speed. To capitalize on this, a number of recent works have naturally tried to combine DP with adaptive optimizers such as Adagrad, RMSProp, and Adam, which have proven to be effective for non-private machine learning tasks, especially those involving sparse gradients or non-uniform stochastic noise~\cite{duchi2011adaptive,hinton2012rmsprop,kingma2014adam,reddi2018adaptive,reddi2019convergence,zhang2019adaptive}.

\begin{figure}[t!]
    \centering
    \includegraphics[width=0.49\textwidth]{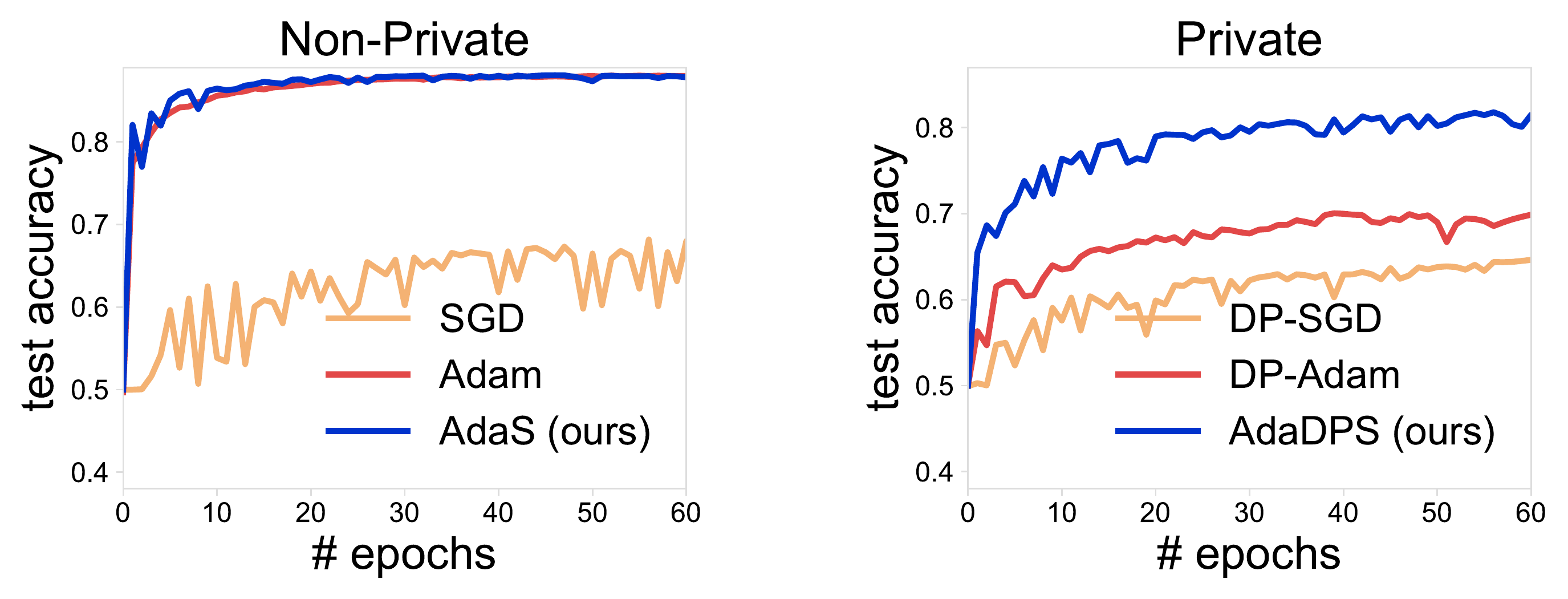}
    \caption{Test performance on IMDB with logistic regression. \texttt{AdaS} refers to using preconditioning in \name for non-private training. Adaptive methods (Adam) become less effective when trained with differential privacy (DP-Adam), while \name retains the benefits of adaptivity.}
    \label{fig: motivation}
\end{figure}

Unfortunately, tasks where adaptive optimizers work particularly well (e.g., sparse, high-dimensional problems), are exactly the tasks where DP is known to degrade performance~\cite{bassily2014private}.
Indeed, as we show in Figure~\ref{fig: motivation}, this can result in existing private adaptive optimization methods performing \textit{only marginally better} than simple baselines such as differentially private stochastic gradient descent (DP-SGD), even under generous privacy budgets.

In this work, we aim to close the gap between adaptive optimization in non-private and private settings. 
We propose \name, a simple yet powerful framework  leveraging non-sensitive side information to effectively adapt to the gradient geometry. Our framework makes two key changes over prior art in private adaptive optimization: (1) Rather than privatizing the gradients first and then applying preconditioners, we show that transforming the gradients prior to privatization can reduce detrimental impacts of noise; (2) To perform gradient transformations, we explore using simple, easily obtainable side information in the data. We discuss two practical scenarios for obtaining such information below.

\paragraph{With Public Data.} A natural choice for side information is to use a small amount of public data generated from a similar distribution as the private data, a common assumption in private optimization~\cite{amid2021public,asi2021private,kairouz2020fast,zhou2020bypassing,zhou2020private}.  In practice, public data could be obtained through proxy data or from ‘opt-out’ users who are willing to share their information~\cite{kairouz2020fast,aldaghri2021feo2}.  Indeed, the notion of heterogeneous DP where subsets of samples require zero or weak privacy has been extensively studied in prior works~\citep[e.g.,][]{alaggan2015heterogeneous,jorgensen2015conservative}. Another line of works is to assume that the gradients are low rank, and then use public data to estimate this gradient subspace---thereby mitigating some of the earlier discussed poor performance of DP in high-dimensional regimes. We do not consider using public data for this purpose in this work, as such a low-rank assumption might not hold in practice, particularly for the problems settings where adaptive optimizers are known to excel~\cite{asi2021private}. Instead, we propose to use the public data more directly: we estimate gradient statistics on public data at each iteration, and then apply these statistics as a preconditioner \textit{before} privatizing the gradients. Despite the simplicity of this procedure, we unaware of any work that has explored it previously.

\paragraph{Without Public Data.} Of course, there may also be applications where it is difficult to obtain public data, particularly data that follows the same distribution as the private data. In such scenarios, our insight is that for many applications, in lieu of public data we may have access to some common knowledge about the training data that can be (i) computed before training, and (ii) used to improve optimization performance in both private and non-private settings. For instance, in many language tasks, certain aggregate statistics (e.g., frequency) of different words/tokens are common knowledge or may be easily computed prior to training, and serve as reasonably good estimates of the predictiveness of each feature. \name considers leveraging such simple heuristics to precondition the gradients. 
Perhaps surprisingly, in our experiments (Section~\ref{sec:exp}), we demonstrate that the performance of \name when scaling gradients via these simple statistics (such as feature frequency) can even match the performance of \name with public data. 

We summarize our main contributions below.
\begin{itemize}[leftmargin=*]
\setlength\itemsep{0em}
\item We propose a simple yet effective framework, \name, to precondition the gradients with non-sensitive side information before privatizing them to realize the full benefits of adaptive methods in private training. Depending on the application at hand, we show that such side information can be estimated from either public data or some common knowledge about the data (e.g. feature frequencies or TF-IDF scores in NLP applications). 
\item We analyze \name and provide convergence guarantees for both convex and non-convex objectives. In convex cases, we analyze a specific form of \name using RMSProp updates to provably demonstrate the benefits of our approach relative to differentially private SGD when the gradients are sparse.
\item Empirically, we evaluate our method on a set of real-world datasets. \name improves the absolute accuracy by 7.7\% on average compared with strong baselines under the same privacy budget, and can even achieve similar accuracy as adaptive methods in non-private settings. We additionally demonstrate how to apply \name to the application of federated learning,
 where it outperforms existing baselines by a large margin.
\end{itemize}

\section{Related Work} \label{sec:related_work}

There are many DP algorithms for machine learning, including object perturbation, gradient perturbation, and model perturbation. In this work, we focus on the popular gradient perturbation method with Gaussian mechanisms~\cite{dwork2014algorithmic}. Without additional assumptions on the problem structure, DP algorithms can suffer from $O(\frac{\sqrt{d}}{n\varepsilon})$ excess empirical risk where $d$ is the dimension of the model parameters and $n$ is the number of training samples~\cite{bassily2014private}. While it is possible to mitigate such a dependence in the unconstrained setting~\cite{kairouz2020fast,song2020evading} or assuming oracle access to a constant-rank gradient subspace~\cite{zhou2020bypassing,kairouz2020fast}, we do not focus on such a setting in this work, as it does not align with the settings in which adaptive methods have been designed~\cite{duchi2011adaptive,asi2021private}. Recent work~\cite{amid2021public} proposes to use public data differently (evaluating public loss as the mirror map in a mirror descent algorithm) to obtain dimension-independent bounds. However, this method do not account for gradient preconditioning, as their approximation is a linear combination of private and public gradients. We empirically verify \name's superior performance in Section~\ref{sec:exp} (Table~\ref{table: compare_PDA-DPMD}).

In the context of adaptive differentially private optimization, we note that `adaptivity' may have various meanings.
For example, \citet{andrew2019differentially} adaptively set the clipping threshold based on the private estimation of gradient norms, which is out of the scope of this work. Instead, our work is related to a line of work that aims to develop and analyze differentially private variants of common adaptive optimizers (e.g., private AdaGrad, private Adam), which mostly focus on estimating gradient statistics from noisy gradients~\cite{zhou2020private,asi2021private,pichapati2019adaclip}. They differ from \name in the order of preconditioning and privatization, the techniques used to approximate the gradient geometry, and the convergence analysis (see Section~\ref{sec:analysis} for details). In empirical evaluation (Section~\ref{sec:exp}), we compare \name with these works on diverse tasks and show that even \name without public data can outperform them.

\section{Preliminaries and Setup}

In terms of privacy formulations, we consider classic sample-level DP in centralized settings (Section~\ref{sec:exp:centralized}), and a variant of it---\textit{user-level DP}---in distributed/federated environments (Section~\ref{sec:exp:fl}). We define both more formally below.
\begin{definition}[Differential privacy~\cite{dwork2006calibrating}]\label{def:dp}
A randomized algorithm $\mathcal{M}$ is $(\varepsilon, \delta)$-differentially private if for all neighbouring datasets $D, D'$ differing by one element, and every possible subset of outputs $O$,
\medmuskip=0mu
\thinmuskip=0mu
\thickmuskip=0mu
\begin{align*}
    \Pr\left(\mathcal{M}(D) \in O\right) \leq e^{\varepsilon} \Pr\left(\mathcal{M}(D') \in O\right) + \delta.
\end{align*}
\end{definition}
Within DP, neighbouring datasets can be defined in different ways depending on the application of interest. In this work, we also apply \name to federated learning (Section~\ref{sec:exp:fl}), where differential privacy is commonly defined at the  granularity of users/devices~\cite{mcmahan2017learning,kairouz2019advances}, as stated below.

\begin{definition}[User-level DP for federated learning~\cite{mcmahan2017learning}]\label{def:user_dp}
A randomized algorithm $\mathcal{M}$ is $(\varepsilon, \delta)$-differentially private if for all datasets $U, U'$ differing by \textit{one user}, and every possible subset of outputs $O$,
\medmuskip=0mu
\thinmuskip=0mu
\thickmuskip=0mu
\begin{align*}
    \Pr\left(\mathcal{M}(U) \in O\right) \leq e^{\varepsilon} \Pr\left(\mathcal{M}(U') \in O\right) + \delta.
\end{align*}
\end{definition}
In centralized empirical risk minimization, our goal is to learn model parameters $w \in \mathbb{R}^d$ to fit $n$ training samples $\{x^i\}_{i\in [n]}$: 
    $\min_w ~F(w) ~=~ \frac{1}{n} \sum_{i=1}^n f(x^i;w)$,
where $f(\cdot)$ is the individual loss function. Optionally, there may exist pubic data denoted as $x_{\text{pub}}$, which does not overlap with $\{x^i\}_{i \in [n]}$. We focus primarily on the classic centralized training, and later on extend our approach to federated settings (Objective~\eqref{obj:fl})~\cite{mcmahan2017communication}.


\begin{figure}[h!]
    \centering
    \includegraphics[trim=10 15 10 10,clip, width=0.27\textwidth]{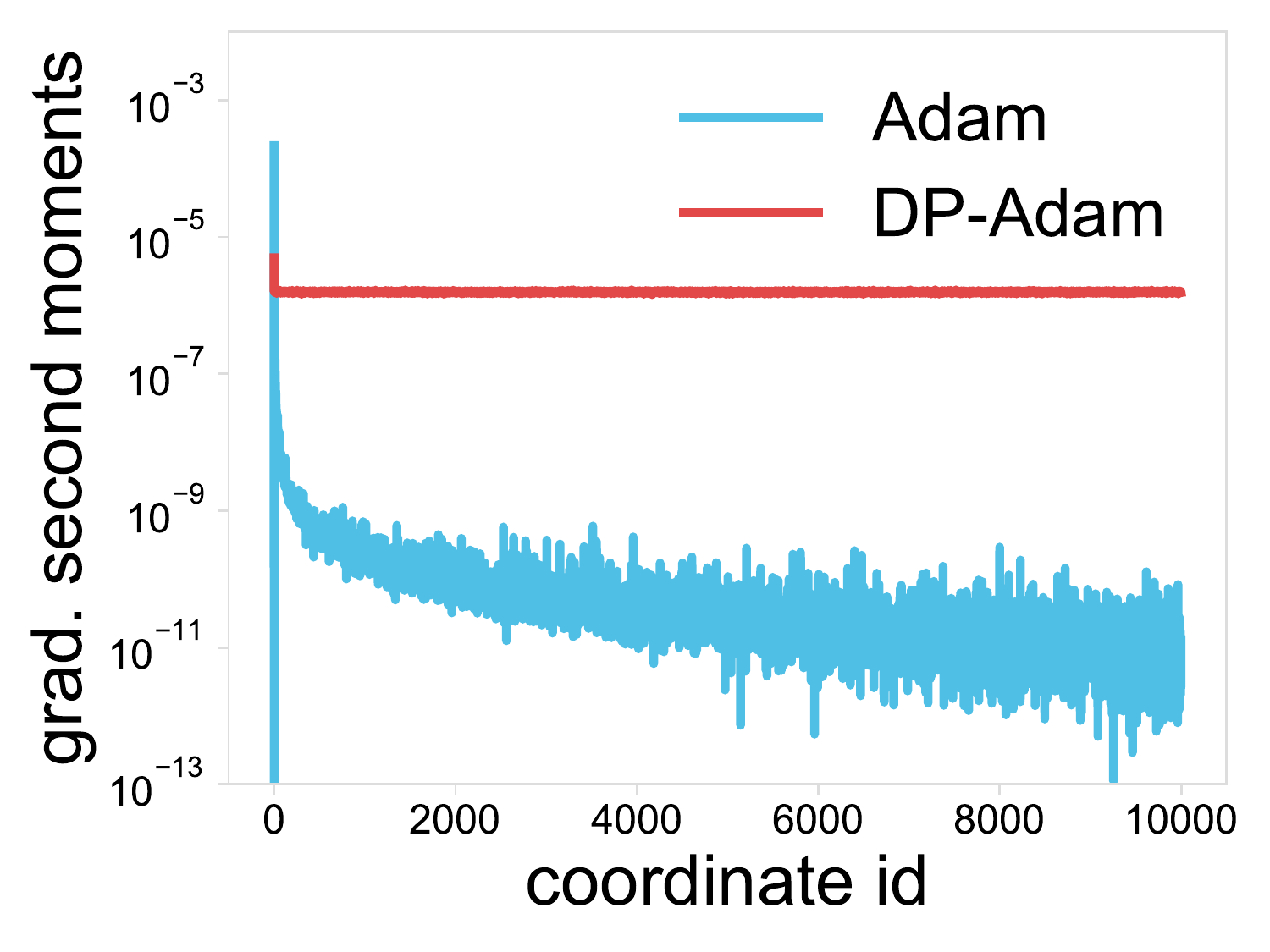}
    \caption{The estimates of gradient statistics (e.g., second moments) in private adaptive methods (e.g., DP-Adam) are noisy and may become uninformative of the relative importance of coordinates.}
    \label{fig: moment_estimation}
\end{figure}

In gradient-based optimization, adaptive optimizers and their properties have been extensively studied~\citep[e.g.,][]{mukkamala2017variants,reddi2018adaptive,duchi2011adaptive,kingma2014adam}. 
They effectively result in coordinate-wise learning rates, which can be advantageous 
for many learning problems.
The preconditioners can be estimated via a moving average of mini-batch gradients (as in,  e.g., Adam) or simply by calculating the sum of gradients so far (AdaGrad). 
In private settings, estimating the required statistics on noisy gradients can introduce significant noise (Figure~\ref{fig: moment_estimation}), making these methods less effective. 
To address this we introduce \name in the next section, and setup some notation below.


\textbf{Notations.} For vectors $u, v \in \mathbb{R}^d$, we use $u+v$ for coordindate-wise addition, and $\frac{u}{v}$ for coordinate-wise division. For a vector $v$ and scalar $a$, $v+a$ denotes adding $a$ to every dimension of $v$.   
For any vector $v$,  $v_j$ always denotes the $j$-th coordinate of $v$. For example, $g^{i,t}_{j}$ refers to the $j$-th coordinate of gradient $g^{i,t}$. We use $\|\cdot\|_M$ to denote the matrix norm defined as $\|\cdot\|_{M} := \sqrt{\langle \cdot, M \cdot\rangle}$ for a symmetric and positive definite matrix $M \in \mathbb{R}^{d\times d}$, or a diagonal matrix with non-negative diagonal entries $M \in \mathbb{R}^d$.

\section{\name: Private Adaptive Optimization with Side Information} \label{sec:method}

Gradient-based private optimization methods usually update the model parameters with noisy gradients at each iteration and then release the private final models~\cite{abadi2016deep}. To control the sensitivity of computing and summing individual gradients from a mini-batch, methods based on the subsampled Gaussian mechanism typically first clip each individual gradient and then add i.i.d. zero-mean Gaussian noise with variance determined by the clipping threshold and the privacy budget. To use adaptivity effectively, in \name, we instead propose first preconditioning the raw gradients with side information estimated either on public data or via some auxiliary knowledge, and then applying the Gaussian mechanism with noise multiplier $\sigma$ on top. 
\name in centralized training is summarized in Algorithm~\ref{alg:dp-side}. Note that \name is a general framework in that it incorporates a set of private adaptive methods. As described in Algorithm~\ref{alg:dp-side}, the functions $\phi, \varphi$, and $\mathcal{A}$ abstract a set of updating rules of different adaptive methods. Next, we describe the algorithm in more detail and instantiate $\phi, \varphi$, and $\mathcal{A}$.

\setlength{\textfloatsep}{10pt}
\begin{algorithm}[t!]
\SetAlgoLined
\DontPrintSemicolon
\SetNoFillComment
\setlength{\abovedisplayskip}{3pt}
\setlength{\belowdisplayskip}{3pt}
\setlength{\abovedisplayshortskip}{3pt}
\setlength{\belowdisplayshortskip}{3pt}
\KwIn{$T$, batch size $b$, noise multiplier $\sigma$, clipping threshold $C$, initial model $w^1 \in \mathbb{R}^d$, side information $A^t \in \mathbb{R}^d$, learning rate $\alpha^t$, potential momentum buffer $M^0 \in \mathbb{R}^d$}
\caption{\name}
\label{alg:dp-side}
    \For{$t=1, \cdots, T-1$}{
        Uniformly sample a mini-batch $B$ ($|B|$=$b$) from the training set and get $b$ gradients:
        \begin{align}
            g^{i,t} \gets \nabla f(x^i; w^t),~ i \in B \nonumber
        \end{align} \;
        \vspace{-2mm}
        \begin{tikzpicture}[remember picture, overlay]
        \draw[line width=0pt, draw=orange!40, rounded corners=2pt, fill=orange!40, fill opacity=0.3]
            ([xshift=0pt,yshift=5pt]$(pic cs:a) + (192pt,6pt)$) rectangle ([xshift=0pt,yshift=0pt]$(pic cs:b)+(-3pt,-86pt)$);
        \end{tikzpicture}
        \textbf{Option 1:} With public data $x_{\text{pub}}$ \\
       ~ Uniformly sample a mini-batch $B'$ ($|B'|$=$b$) from $x_{\text{pub}}$, get gradients, and update $A^t$ and $M^t$ with recurrence $\phi$ and $\varphi$ respectively:
        \begin{align*} 
            &\hat{g}^t \gets \frac{1}{b} \sum_{j \in B'} \nabla f(x^j; w^t), ~x^j \in x_{\text{pub}} \\
            &A^t \gets \phi(A^{t-1}, \hat{g}^t), M^t \gets \varphi(M^{t-1}, \hat{g}^t), 
        \end{align*}  \\
        \begin{tikzpicture}[remember picture, overlay]
        \draw[line width=0pt, draw=blue!30, rounded corners=2pt, fill=blue!30, fill opacity=0.3]
            ([xshift=0pt,yshift=2pt]$(pic cs:a) + (192pt,7pt)$) rectangle ([xshift=2pt,yshift=2pt]$(pic cs:b)+(-5pt,-23pt)$);
        \end{tikzpicture}
        \vspace{0mm}
        \textbf{Option 2:} Without public data
        \begin{align*}
            A^t \text{ estimated via heuristics} 
        \end{align*} \\
        Precondition individual gradients by $\mathcal{A}$:
        \begin{align*}
            g^{i,t} \gets \mathcal{A}(g^{i,t}, A^t, M^t)
        \end{align*} \\
        Privatize preconditioned gradients: 
        \begin{align}
            \Tilde{g}^t \gets \frac{1}{b} \sum_{i \in B} \text{clip}\left(g^{i,t}, C\right) + \frac{1}{b} \mathcal{N}\left(0, \sigma^2 C^2\right),\nonumber
        \end{align}
        where $\text{clip}(g, C)$ clips a vector $g$ to $L_2$ norm $C$\;
        Update the model parameter $w$ as
        \begin{align*}
            w^{t+1} \gets w^t - \alpha^t \Tilde{g}^t
        \end{align*}
        }
    \Return{$w^T$} \;
\end{algorithm}

{\color{myorange} \textbf{Option 1 (With Public Data).}} 
We first consider estimating gradient statistics based on public data. Functions $\phi, \varphi$, and $\mathcal{A}$ can define a very broad a set of common adaptive updating rules, as shown below.

\begin{itemize}[leftmargin=*]
\setlength\itemsep{0em}
    \item \textit{Adam}: $A^t$ is the square root of the second moment estimation, and $M^t$ is the first moment estimation; with $\mathcal{A}(g^{i,t}, A^t, M^t)=\frac{\beta^t g^{i,t}+(1-\beta^t) M^t}{A^t}$ for some moving averaging parameter $\beta^t$. 
    \item \textit{AdaGrad}: The update corresponds to $M^t = \mathbf{0}$, $(A^t)^2 = (A^{t-1})^2+(\hat{g}^t)^2$, and $\mathcal{A}(g^{i,t}, A^t, M^t)=\frac{ g^{i,t}}{A^t}$.
    \item \textit{RMSProp}: $A^t$ is the square root of the second moment estimation with $M^t=\mathbf{0}$. And  $\mathcal{A}(g^{i,t}, A^t, M^t)=\frac{ g^{i,t}}{A^t}$.
\end{itemize}
We note that the \name framework can potentially incorporate other adaptive optimizers, beyond what is listed above. In our analysis and experiments, we mainly focus on using RMSProp updates to obtain the preconditioner, because AdaGrad which sums up gradients in all iterations so far in the denominator, often has poor practical performance, and Adam needs to maintain an additional mean estimator. However, we also evaluate the use of Adam within \name in Table~\ref{table: all_methods} in Appendix~\ref{app:exp_additional}, showing that it yields similar improvements as RMSProp across all datasets. 
In Section~\ref{sec:analysis}, we analyze the convergence of $A^t$ as $\mathbb{E}[(\hat{g}^t)^2]$, and prove that in sparse settings, \name allows the  addition of  less noise under the same privacy budget. 

\textcolor{myblue}{\textbf{Option 2 (Without Public Data).}}  When there is no public data available, we develop simple and effective heuristics to determine which coordinates are more predictive based on non-sensitive side information.  In particular, for generalized linear models in NLP applications, we set $A^t$ to be (i) proportional to the frequency of input tokens, or (ii) proportional to the TF-IDF values of input tokens. Follow a similar analysis as that of Option 1, we provide theoretical justification in Theorem~\ref{thm:fixed_A} in Section~\ref{sec:analysis:convex:linear}. While these are simple approaches to remove the dependence on public data, we find that they can significantly outperform DP-SGD for real-world tasks with several million model parameters (Section~\ref{sec:exp:centralized}).

\textbf{Privacy guarantees.} We now state the differential privacy guarantees of Algorithm~\ref{alg:dp-side}. As the side information $A^t$ (as well as the potential momentum buffer $M^t$) is non-sensitive, its privacy property directly follows from previous results~\cite{abadi2016deep}. 
\begin{theorem}[Privacy guarantee of Algorithm~\ref{alg:dp-side}~\cite{abadi2016deep}] 
Assume the side information $A^t$ is non-sensitive. There exist constants $c_1$ and $c_2$ such that for any $\varepsilon < c_1 b^2 T/n^2$, Algorithm~\ref{alg:dp-side} is $(\varepsilon, \delta)$-differentially private for any $\delta > 0$ if $\sigma \geq c_2 \frac{b\sqrt{T \log(1/\delta)}}{n\varepsilon}$.
\end{theorem}

\subsection{Intuition for $A^t$} 
In this section we provide further intuition for the \name framework. 
When $A^t$ is an all-ones vector, \name reduces to the normal DP-SGD algorithm. Otherwise, $A^t$ is indicative of how informative each coordinate is. Intuitively, suppose clipping does not happen and the public data come from the same distribution as private data so that for the RMSProp preconditioner, we have $A^t=\sqrt{\mathbb{E}[(g^{i,t})^2]}$. Then the effective transformation on each individual gradient $g^{i,t}$ is $ \frac{g^{i,t} + \mathcal{N}\left({0},\sigma^2 C^2 \mathbb{E}[(g^{i,t})^2]\right)}{\sqrt{\mathbb{E}[(g^{i,t})^2]}}$.
This can be viewed as first adding non-isotropic noise proportional to the second moment of gradients, and then applying RMSProp updates, which is beneficial as  coordinates with higher second moments are more tolerant to noise. Therefore, \name could improve privacy/utility tradeoffs via adding coordinate-specific noise (formalized in Theorem~\ref{thm: convex_convergence}).

\begin{figure}[h!]
    \centering
    \includegraphics[width=0.5\textwidth]{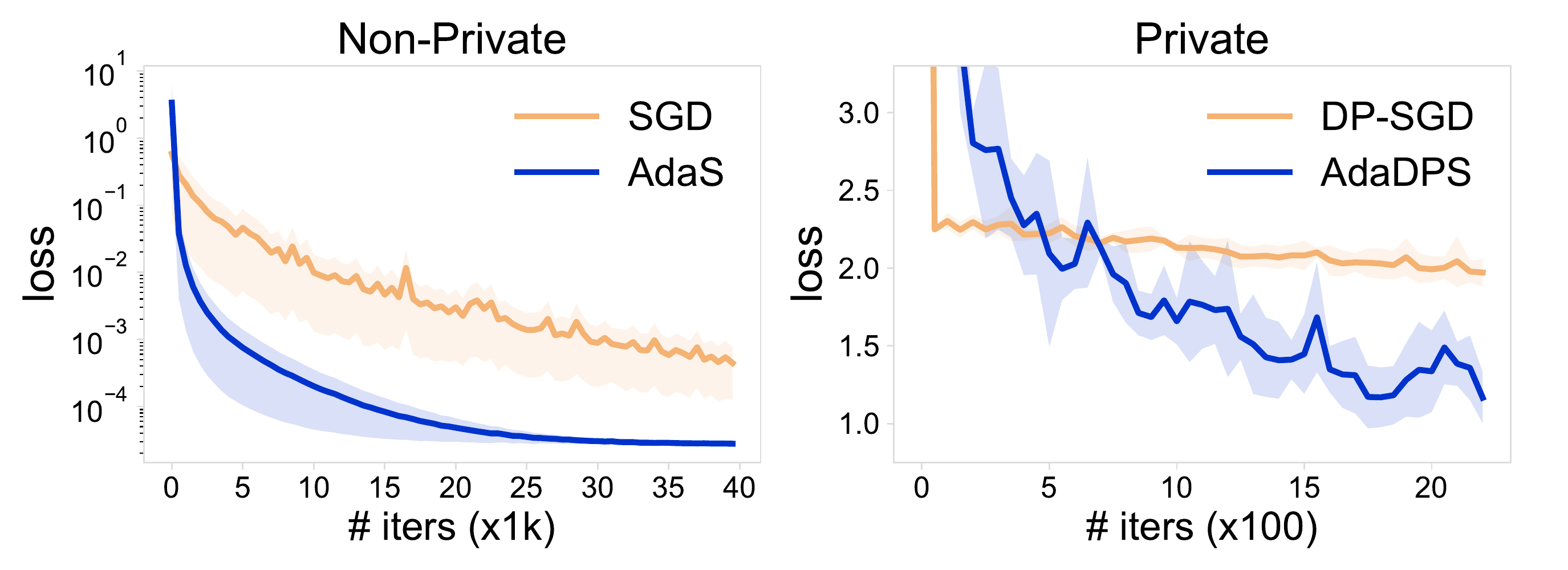}
    \caption{Training loss on the linear regression problem described in Section~\ref{sec:method} (averaged over five runs). We tune optimal learning rates separately for each method. Private training (right) achieves $(4.13,10^{-3})$-DP.}
    \label{fig:toy}
\end{figure}

We next consider a toy example to highlight one of the regimes where \name (or, adaptive methods) is particularly effective. Consider a linear regression task with the objective $\min_{w \in \mathbb{R}^{500}} \frac{1}{2n}\sum_{i \in [n]} (w^\top x^i-y^i)^2$ where  $n=1,000$ and each sample $x^i \in \mathbb{R}^{500}$. 
In many real-world applications, 
the tokens (features) are sparse and their frequencies follow heavy-tailed distributions.  Without loss of generality, we assume the first 10\% features are frequent and uninformative; and the later 90\% rare and informative. Let the $j$-th feature of all data points be sampled from a random variable $x_j \in \{0,1\}$. Features and the underlying true $w$ are generated as follows:
\medmuskip=0mu
\thinmuskip=0mu
\thickmuskip=0mu
\begin{align*}
    \Pr(x_{j}=1) = \begin{cases}
                    0.9, &j \leq 50 \\
                    0.01, &j > 50 
                    \end{cases},
    \quad w_j = \begin{cases}
            0.01, &  j \leq 50 \\
            1.0, & j > 50
            \end{cases}.
\end{align*}
Labels are generated by $y^i = \left\langle w, x^i \right\rangle  + b^i$ where $b^i \sim \mathcal{N}(0, 0.01)$. For \name, we assume model engineers know which words are more frequent, thus setting $A_j^t=1$ for $j \leq 50$ and $A_j^t=0.01$ otherwise for all $t$. Using larger learning rates on informative coordinates, side information helps to improve optimization performance dramatically  (see results in Figure~\ref{fig:toy}). 

\textbf{Comparison to~\citet{asi2021private}.} The most related work to ours is~\citet{asi2021private}, which adds non-isotropic noise which lies in a non-uniform ellipsoid to the gradients, and (optionally) transforms the resulting gradients with the demoninator used in AdaGrad. \name differs from their approach in several ways, as we (i) first precondition then add noise (as opposed to the other way round), (ii) consider a broader class of preconditioners (beyond AdaGrad), and (iii) make the approaches to estimating gradient geometry in~\citet{asi2021private} more explicit in lieu of public data (as discussed in previous sections). 
Empirically, we compare with another state-of-the-art method~\cite{amid2021public} which outperforms~\citet{asi2021private}, and demonstrate \name's superior performance (Section~\ref{sec:exp}, Table~\ref{table: compare_PDA-DPMD}).\footnote{We do not compare with~\citet{asi2021private} directly as the code is not publicly available.}

\section{Convergence Analysis} \label{sec:analysis}

We now analyze the convergence of \name (Algorithm~\ref{alg:dp-side}) with and without public data, in both convex (Section~\ref{sec:analysis:convex}) and non-convex (Section~\ref{sec:analysis:non-convex}) settings. When there is public data available, we prove that \name adds less noise (plus, with noise proportional to the magnitude of gradients) compared with DP-SGD. Our theory extends previous proofs in related contexts, but considers stochastic gradients, adding random Gaussian noise to the processed gradients, and estimating the preconditioner on public data (as opposed to updating it with the raw gradients on training data). When there is no public data, we present convergence results for general $A^t$, covering the heuristics used in practice.

\subsection{Convex Convergence} \label{sec:analysis:convex}
For convex functions, we define the optimal model $w^*$ as
    $w^* \in~\arg\min_w F(w)$.
First we state a set of assumptions that are used in the analyses.
\begin{assumption} \label{assum:bounded_domain}
There exists a constant $D$ such that $\|w^t-w^*\|_2 \leq D$ for any $t \in [T]$.
\end{assumption}
\begin{assumption}\label{assum:bounded_sensitivity}
There exists a constant $C$ such that $\left\|\frac{g^{i,t}}{A^t}\right\|_2 \leq C$ for any $t \in [T]$ and $i \in [n]$.
\end{assumption}
\begin{assumption}\label{assum:bounded_private_public_gap} 
Denote $g^t := \frac{1}{b} \sum_{i \in B} g^{i,t}$. There exists a constant $a \in (0,1]$ such that for any $j \in [d]$ and $t \in [T]$, $a (\hat{g}_j^t)^2 \leq (g_j^t)^2 \leq \frac{1}{a}(\hat{g}_j^t)^2$ holds. 
\end{assumption}
Assumption~\ref{assum:bounded_sensitivity} aims to bound the $L_2$ norm of the transformed gradient, thus resulting in bounded $L_2$ sensitivity on the operation of calculating and averaging (scaled) individual gradients from a mini-batch. Assuming bounded stochastic gradient norm is standard in prior works on convex and non-convex private optimization~\citep[e.g.,][]{kairouz2020fast,zhou2020private}.  Assumption~\ref{assum:bounded_private_public_gap} bounds the dissimilarity between public and private data. 

Within the framework of \name, we explore the convergence of the RMSProp preconditioner where $A^t = \sqrt{\mathbb{E}[(\hat{g}^t)^2]}+\epsilon^t$ (with public data) where $\epsilon^t$ is some small constant or $A^t$ is obtained via side information heuristics. We first look at the case where there exist public data, and therefore the exact updating rule at the $t$-th iteration is:
\medmuskip=0mu
\thinmuskip=0mu
\thickmuskip=0mu
\begin{align*}
     \hat{g}^t, g^t \gets \frac{1}{b} \sum_{j \in B'} \nabla f(x^j; w^t) ~(x^j \in x_{\text{pub}}), \frac{1}{b} \sum_{i \in B} \nabla f(x^i;w^t), 
\end{align*}
\vspace{-1em}
\begin{align*}
    v^t \gets \beta^t v^{t-1} + (1-\beta^t) (\hat{g}^t)^2, ~ A^t \gets\sqrt{v^t}+\epsilon^t, 
\end{align*}
\vspace{-1em}
\begin{align*}
    w^{t+1} \gets w^t - \alpha^t \left(\frac{g^t}{A^t}+N\right), ~N\sim \frac{1}{b}\mathcal{N} ({0}, \sigma^2 C^2).
\end{align*}
Theorem~\ref{thm: convex_convergence} below states the convergence guarantees.
\begin{theorem} \label{thm: convex_convergence}
Assume $F(w)$ is a convex function w.r.t. $w$. Let Assumptions~\ref{assum:bounded_domain}-\ref{assum:bounded_private_public_gap} hold. Additionally, choose $\beta^t$ such that $1-\frac{\gamma}{t} \geq \beta^t \geq 1-\frac{1}{t}$ holds for some $\gamma \in (0,1]$; and let $\sqrt{t+1} \epsilon^{t+1} \geq \sqrt{t}\epsilon^t$ for any $t$. After running Algorithm~\ref{alg:dp-side} using learning rates $\alpha^t=\frac{\alpha}{\sqrt{t}}$ with public data for $T$  iterations, we have
\medmuskip=0mu
\thinmuskip=0mu
\thickmuskip=0mu
\begin{align*}
    \min_{t \in [T]} \mathbb{E}[F(w^t)] - F^* \leq  \frac{G}{\sqrt{T}} \sum_{j=1}^d \mathbb{E}\left[A_j^T \right]+\frac{\alpha}{\sqrt{T}}\max_{t\in [T]} \mathbb{E}\left[\|N\|^2_{A^t}\right],
\end{align*}
where $F^* := F(w^*)$, $G = \frac{D^2}{2\alpha} + \frac{\alpha (2-\gamma)}{a\gamma}$, and $A_j^T=\sqrt{v^T_j} +  \epsilon^T$.
\end{theorem}
The full proof is deferred to Appendix~\ref{app:convex_proof}. Our proof extend the proof of the original RMSProp method with full gradients in~\citet{mukkamala2017variants} to stochastic and private settings. 
From Theorem~\ref{fig:with-public-private}, we see that the first term in the bound is standard for the RMSProp optimizer, and the last term is due to noise added to ensure differential privacy. Fixing the noise multiplier $\sigma$, the second term depends on the clipping value $C$ and the preconditioner $A^t$. We see that when the gradients are sparse, it is likely that the added DP noise would be significantly reduced. In other words, to guarantee overall $(\varepsilon, \delta)$-differential privacy by running $T$ total iterations, we can set $\sigma^2 = O\left(\frac{b^2  T \log(1/\delta)}{n^2 \varepsilon^2}\right)$ and $T=O\left(\frac{n^2\varepsilon^2}{ \max_{t\in [T]} \mathbb{E}\left[\left\|A^t\right\|_1\right] \log(1/\delta)}\right)$.  The convergence rate therefore becomes
\begin{align*}
    \min_{t\in [T]}\mathbb{E}[F(w^t)] - F^* \leq O\left(\frac{\sqrt{\max_{t\in [T]} \mathbb{E}\left[\left\|A^t\right\|_1\right]}}{ n\varepsilon}\right).
\end{align*}
When gradients are sparse (hence $\max_{t\in [T]} \mathbb{E}\left[\left\|A^t\right\|_1\right] < d$), the amount of noise added will be significantly smaller compared with that of vanilla DP-SGD to guarantee the same level of privacy. This highlights one regime where \name is particularly useful, though \name also yield improvements in other settings with dense gradients (Table~\ref{table: all_methods} in the appendix). Here, Theorem~\ref{thm: convex_convergence} assumes access to $\mathbb{E}[(\hat{g}^t)^2]$. When there is no public data available, we leverage other easily obtainable side information to determine fixed $A^t$ prior to training, as analyzed in the next section. 

\subsubsection{Fixed $A^t$} \label{sec:analysis:convex:linear}
\begin{theorem} \label{thm:fixed_A}
Let assumptions in Theorem~\ref{thm: convex_convergence} hold. Running Algorithm~\ref{alg:dp-side} using learning rates $\alpha^t=\frac{\alpha}{\sqrt{t}}$ without public data with side information $A \in \mathbb{R}^d$ for $T$ iterations gives
\begin{align*}
     &\min_{t \in [T]} \mathbb{E}[F(w^t)] - F^*  \leq O\left(\frac{\alpha R+1}{\sqrt{T}} \sum_{j=1}^d {A}_j + \frac{\alpha}{\sqrt{T}} \mathbb{E}\left[\|N\|^2_{{A}}\right]\right),
\end{align*}
where $R := \max_{j,t} \frac{\mathbb{E}[(g_j^t)^2]}{A_j^2}$ and $g^t := \frac{1}{b} \sum_{i \in B} g^{i,t}$.
\end{theorem} 

From Theorem~\ref{thm:fixed_A} above, we observe that ${A}$ should be chosen such that both $R$ and $\sum_{i=j}^d {A}_j$ are minimized. For a large class of generalized linear models, we are able to obtain appropriate $A$ based on the feature space information to control the values of $R$, as discussed in the following.

Considering generalized linear models. Under the model parameter $w \in \mathbb{R}^d$, for any $x^i$, the gradients are $c(x^i; w) x^i$ where $c(x^i;w) \in \mathbb{R}$ is a function of $x^i$ and $w$. We assume that for any $i \in [n]$ and $w$, there exists a constant $c_{\max}$ such that
    $|c(x^i;w)| \leq c_{\max}$. 
One natural choice of $A$ is to set ${A}_j = \sqrt{\mathbb{E}[x_j^2]}+\epsilon$ for each coordinate $j \in [d]$ (such that $R\leq c_{\max}^2$), which could improve the noise term $\mathbb{E}\left[\|N\|^2_A\right]$ when the features are sparse. Nevertheless, $\mathbb{E}[x^2]$ can be unrealistic to obtain prior to training. Instead, one side information of choice is $\mathbb{E}[x]$, which implies how rare the raw features are in some NLP applications.
Let ${A}_j = \mathbb{E}[|x_j|]+\epsilon~(j \in [d])$. 
Then
\begin{align*}
    R = \max_{j, t} \frac{\mathbb{E}\left[(g_j^t)^2\right]}{\left(\mathbb{E}[|x_j|]+\epsilon\right)^2} \leq 
    \max_{j, t} \frac{c_{\max}^2 \mathbb{E}[x_j^2]}{(\mathbb{E}[|x_j|])^2 + \epsilon^2}.
\end{align*}
To reason about how large $R$ is, we consider a simple setup where each feature takes the value of $v>0$ with probability $p$, and $0$ with probability $1-p$.  
It is straightforward to see the scaling of the last two terms in the convergence bound in Theorem~\ref{thm:fixed_A}:
\medmuskip=0mu
\thinmuskip=0mu
\thickmuskip=0mu
\begin{align*}
    R \sum_{j=1}^d {A}_j &= O\left(\frac{1}{p}\right) O\left(d pv\right) = O(dv), \\
    \mathbb{E}\left[\|N\|^2_{{A}}\right] &=  \sigma^2 C^2 \sum_{j=1}^d {A}_j = O\left(\max_{i,t} \|g^{i,t}\|^2 \frac{dpv}{(pv+\epsilon)^2}\right).
\end{align*}
$\mathbb{E}\left[\|N\|_A^2\right]$ will reduce to $O(d)$ if the gradients are sparse in a certain way, i.e., $\max_{i,t} \|g^{i,t}\|^2 = O(p)$. Note that only using this simple first moment information ($A_j=\mathbb{E}\left[|x_j|\right]+\epsilon$), we are not able to obtain a constant improvement in the convergence bound as in Theorem~\ref{thm: convex_convergence} with public data. However, we empirically demonstrate that these ideas can be effective in practice in Section~\ref{sec:exp}.

\subsection{Non-Convex Convergence} \label{sec:analysis:non-convex}
In addition to convex problems, we also consider convergence to a stationary point for non-convex and smooth functions. We introduce additional assumptions in the following.
\begin{assumption} \label{assum:smoothness}
    $F(\cdot)$ is $L$-smooth.
\end{assumption}
\begin{assumption} \label{assum:bounded_variance}
    The expectation of stochastic gradient variance is bounded, i.e., $\mathbb{E}[\|g_j^{i,t}-\mathbb{E}[g_j^{i,t}]\|_2^2] \leq \tau_j^2$ for all $i,t,j$. Denote $\tau^2 := (\tau_1^2,\cdots,\tau_d^2) \in \mathbb{R}^d$.
\end{assumption}
Assumption~\ref{assum:smoothness} together with Assumption~\ref{assum:bounded_domain} implies that there exists a constant that bounds $\|\nabla F(w^t)\|$ for any $t$, which we denote as $B$.
\begin{theorem} \label{thm:convergence_non_convex}
Let Assumptions~\ref{assum:bounded_domain}-\ref{assum:bounded_variance} hold. After running Algorithm~\ref{alg:dp-side} with public data for $T$ iterations using a constant learning rate $\alpha$ and a constant $\epsilon$, choosing the constants to satisfy $\alpha \leq \frac{\epsilon}{2L}$ and $B \sqrt{1-\beta} \leq \frac{\sqrt{a}\epsilon}{4}$, we have
\medmuskip=0mu
\thinmuskip=0mu
\thickmuskip=0mu
\begin{align*}
    \frac{1}{T} \sum_{t\in[T]} \mathbb{E}\left[ \|\nabla F(w^t)\|^2\right] \leq O\left(\frac{1}{T}\right) + O\left(\frac{\|\tau^2\|_1}{b} + \frac{\sigma^2}{b^2}\right). 
\end{align*}
\end{theorem}
The proof is mostly extended from Adam's proof in~\citet{reddi2018adaptive} (see Appendix~\ref{app:non-convex_proof} for complete steps). When the batch size increases, both the stochastic gradient noise and differential privacy noise would be reduced.
\begin{theorem} \label{thm:non-convex_fix_A}
Let Assumptions~\ref{assum:bounded_domain}-\ref{assum:bounded_variance} hold. After running Algorithm~\ref{alg:dp-side} with a fixed $A$ as prior information  for $T$ iterations using a constant learning rate $\alpha$ and a constant $\epsilon$, choosing the constants to satisfy $\alpha \leq \frac{\epsilon}{L}$, we have
\medmuskip=0mu
\thinmuskip=0mu
\thickmuskip=0mu
\begin{align}
    \frac{1}{T} \sum_{t\in[T]} \mathbb{E}\left[ \|\nabla F(w^t)\|_{A^{-1}}^2\right] \leq O\left(\frac{1}{T}\right) + O\left(\frac{\tau^2}{A}\frac{1}{b} + \frac{\sigma^2}{b^2}\right). \nonumber
\end{align}
\end{theorem}

\section{Experiments} \label{sec:exp}

We evaluate the performance of \name in both centralized (Section~\ref{sec:exp:centralized}) and federated (Section~\ref{sec:exp:fl}) settings for various tasks and models. In centralized training, we investigate two practical scenarios for obtaining side information with and without public data (Section~\ref{sec:exp:centralized:with_public} and~\ref{sec:exp:centralized:without_public}).  We describe our experimental setup below; details of datasets, models, and hyperparameter tuning are described in Appendix~\ref{app:exp_details}. Our code is publicly available at~\href{https://github.com/litian96/AdaDPS}{\texttt{github.com/litian96/AdaDPS}}.

\textbf{Datasets.} We consider common benchmarks for adaptive optimization in centralized or federated settings~\cite{amid2021public,reddi2018adaptive,reddi2020adaptive} involving varying types of models (both convex and non-convex) and data (both text and image data). Both linear and non-convex models contain millions of learnable parameters.

\textbf{Hyperparameters.}  We fix the noise multiplier $\sigma$ for each task, and select an individual (fixed) clipping threshold for each differentially private run. To track the privacy loss (to ensure $(\varepsilon, \delta)$-DP), we add the same amount of noise to all compared methods, set the $\delta$ value to be the inverse of the number of all training samples, and compute $\varepsilon$ using R{\'e}nyi differential privacy (RDP) accountant for the subsampled Gaussian mechanism~\cite{mironov2019r}.

\subsection{Centralized Training} \label{sec:exp:centralized}

\textbf{Common Baselines.} One can directly privatize an adaptive optimizer by \textit{first privatizing} the raw gradients, and \textit{then applying that adaptive method} on top of noisy gradients~\cite{zhou2020private}. 
We consider these baselines named DP-Adam or DP-RMSProp where the adaptive optimizer is chosen to be Adam or RMSProp (same as DP-Adam appearing in previous sections). As the empirical results of DP-Adam and DP-RMSProp are very similar (Table~\ref{table: all_methods} in the appendix), in the main text, we mainly compare \name with DP-Adam~\cite{zhou2020private} and DP-SGD~\cite{abadi2016deep}. For completeness, we present the exact DP-Adam algorithm in Appendix~\ref{app:exp_details}.

\subsubsection{With public data}\label{sec:exp:centralized:with_public}

In the main text, we set the public data size to be 1\% of training data size. We further explore the effects of public data size empirically in Table~\ref{table: public-data-size}, Appendix~\ref{app:exp_additional}. Next, we present results of comparing \name with several baselines, and results using both in-distribution (ID) and out-of-distribution (OOD) data as public data to estimate the preconditioner.

\textbf{Preconditioning \textit{Noisy} Gradients with Public Data.} In addition to DP-SGD and DP-Adam mentioned in Section~\ref{sec:exp:centralized}, we consider another method of preconditioning the \textit{noisy} gradients with second moment estimates obtained from \textit{public data}.  Specifically, the updating rule at iteration $t$ is
\medmuskip=0mu
\thinmuskip=0mu
\thickmuskip=0mu
\begin{align*}
    \Tilde{g}^t &\gets \frac{1}{b} \sum_{i \in B} \text{clip}\left(g^{i,t}, C\right) + \frac{1}{b} \mathcal{N}({0},\sigma^2 C^2),\\
    \Tilde{g}^t &\gets \frac{\Tilde{g}^t}{\sqrt{\mathbb{E} [(\hat{g}^t)^2]} + \epsilon^t}, 
    \text{ where } \hat{g}^t := \nabla f(x; w^t) \text{ for }x \in x_{\text{pub}}. 
\end{align*}
We call this adaptive baseline DP-R-Pub, which is equivalent to standard DP-RMSProp but using public data to estimate the preconditioner. Comparing with this method directly reflects the importance of the order of preconditioning and privatization in \name.

Results with in-distribution proxy data (randomly sampled from training sets) are shown in Figure~\ref{fig:with-public-private} and Table~\ref{table: mnist_ae} below. 
We see that across three datasets, (i) DP-Adam does not necessarily outperform DP-SGD all the same, (ii) \name improves over all baselines significantly, including DP-R-Pub. Full results involving DP-RMSProp and \name with Adam as the updating rule are presented in Table~\ref{table: all_methods}. 

\begin{figure}[h!]
    \centering
    \includegraphics[width=0.49\textwidth]{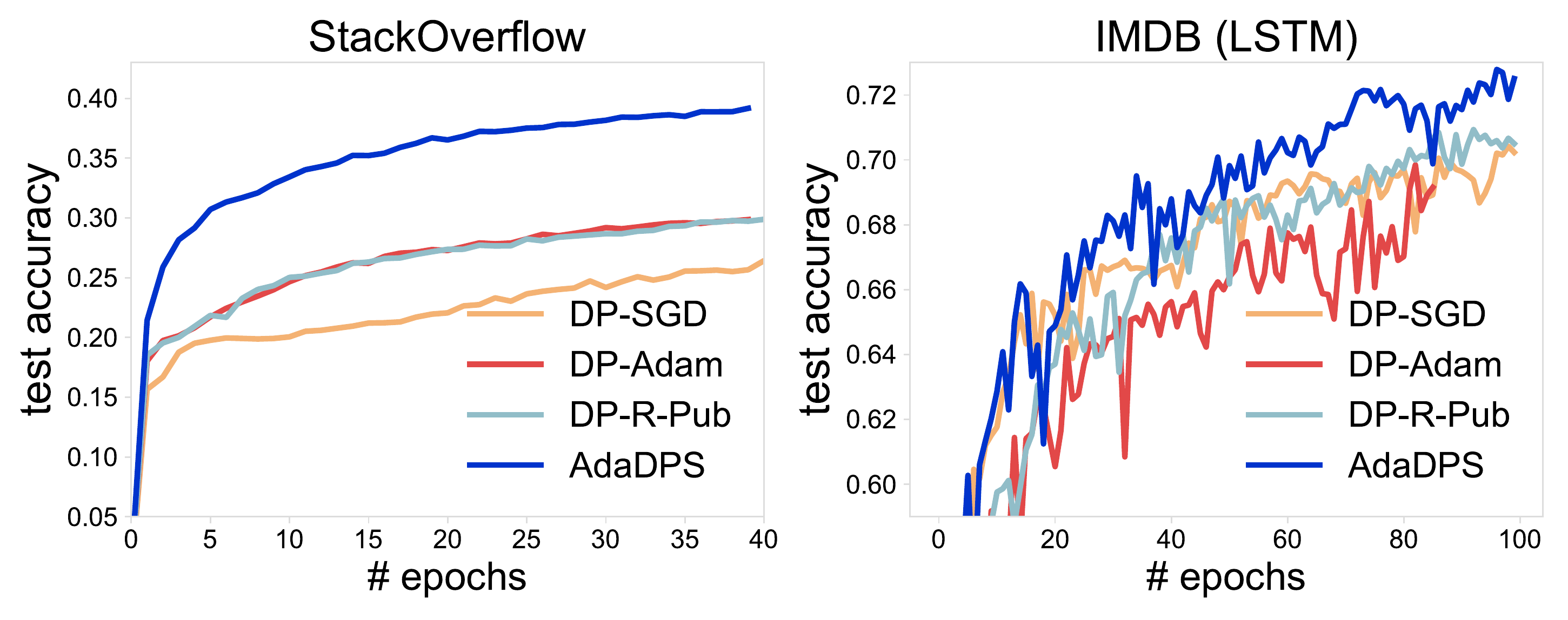}
    \caption{Test accuracies of baselines and \name assuming access to public data. $\varepsilon$ values on these two datasets are 0.84 and 2.8, respectively. \name significantly improves test performance, even reaching an accuracy much higher than the accuracy of SGD in non-private training on StackOverflow.}
    \label{fig:with-public-private}
\end{figure}

\begin{table}[h!]
    \centering
    \scalebox{0.99}{
    \begin{tabular}{@{}l  c  c @{}}
        \toprule[\heavyrulewidth]
       \textbf{Methods} & \textbf{Loss} {\small $\times$100} ($\sigma$=1) & \textbf{Loss} {\small $\times$100} ($\sigma$=0.75)\\
        \hline 
        DP-SGD & 5.013 {\small (.001)} &  3.792 {\small(.001)} \\
        DP-Adam & 3.697 {\small (.020)}  &  3.286 {\small (.016)} \\
        \name{} & \textbf{3.566} {\small (.008)} &  \textbf{3.158} {\small (.003)} \\
        \bottomrule
    \end{tabular}}
    \caption{\small Test reconstruction loss (mean and standard deviation across three runs) on MNIST under a deep autoencoder model. $\sigma$=1 and $\sigma$=0.75 correspond to $\varepsilon$=1.6 and $\varepsilon$=3, respectively. DP-Adam works well in this task compared with DP-SGD. \name improves over DP-Adam.}
    \label{table: mnist_ae}
\end{table}

For completeness, we also evaluate \name on the MNIST image classification benchmark, and observe that it yields $0.5\%-2\%$ accuracy improvements (Table~\ref{table: all_methods}), depending on which specific adaptive method to use.

\paragraph{Comparisons with~\citet{amid2021public}.}
We compare \name with one recent work, PDA-DPMD, which is the state-of-the-art that leverages public data to improve privacy/utility tradeoffs in a mirror descent framework~\cite{amid2021public}. We take their proposed approximation, where the actual gradients are a convex combination of gradients on private and public data. As this approximation does not precondition gradients, PDA-DPMD could underperform \name in the tasks where adaptivity is critical, as shown in Table~\ref{table: compare_PDA-DPMD} below.

\begin{table}[h!]
    \centering
    \scalebox{0.84}{
    \begin{tabular}{@{} l  l c c c @{}}
    \toprule[\heavyrulewidth]
    {\multirow{2}{*}{\textbf{Datasets}}} 
    & \multicolumn{1}{l}{\multirow{2}{*}{
       \textbf{Metrics}}} & \textbf{PDA-DPMD} & \textbf{\name} & \textbf{\name}   \\
               & & w/ public & w/ public & w/o public \\
       \hline
        IMDB & accuracy & 0.62 & \textbf{0.80} & 0.75 \\
        StackOverflow & accuracy & 0.33 & \textbf{0.40} & \textbf{0.41} \\
        MNIST & loss & 0.039 & \textbf{0.036} & --- \\
    \bottomrule
    \end{tabular}}
    \caption{Comparison with a recent method (PDA-DPMD) using public data in private mirror descent. 
    \name outperforms PDA-DPMD due to preconditioning. 
    }
    \label{table: compare_PDA-DPMD}
\end{table}

\paragraph{Out-Of-Distribution Public Data}

As mentioned in Section~\ref{sec:intro}, public data could be obtained via a small amount of proxy data or `opt-out' users that do not need privacy protection. We consider two practical cases where we use OOD data to extract side information. For IMDB sentiment analysis, we use a small subset of Amazon reviews\footnote{\href{https://figshare.com/articles/dataset/Amazon_Reviews_Full/13232537/1}{figshare.com/articles/dataset/Amazon\_Reviews\_Full/13232537/1}}~\cite{zhang2015character} as public data (1\% the size of IMDB). Amazon reviews study a more fine-grained 5-class classification problem on product reviews, and we map labels \{1, 2\} to 0 (negative), and labels \{4, 5\} to 1 (positive). For StackOverflow tag prediction task which consists of 400 users  with different styles and interested topics, we simply hold out the first four of them to provide public data.
We show results in Table~\ref{table: OOD-public-data} below. We see that the preconditioners obtained from out-of-distribution but related public data are fairly effective.

\begin{table}[h!]
    \centering
    \scalebox{0.99}{
    \begin{tabular}{@{} l  c c c @{}}
    \toprule[\heavyrulewidth]
    {\multirow{2}{*}{\textbf{Datasets}}} 
     & \multicolumn{1}{l}{\multirow{2}{*}{
       \textbf{DP-SGD}}} & \textbf{\name}  & \textbf{\name}    \\
              & & OOD public & ID public   \\
       \hline
        IMDB  &  0.63    &  \textbf{0.79}   & \textbf{0.80}  \\
        StackOverflow   &  0.28    &  \textbf{0.40}   & \textbf{0.40}  \\
    \bottomrule
    \end{tabular}}
    \caption{Using small  out-of-distribution data as public data achieves the same improvements. For IMDB, we leverage Amazon reviews data (1\% the size of IMDB) as public data. For StackOverflow, we hold out 1\% users as those who opt out of privacy training. 
    }
    \label{table: OOD-public-data}
\end{table}

\subsubsection{Without public data}\label{sec:exp:centralized:without_public}
When it is difficult to obtain public data that follow sufficiently similar distribution as private training data, we explore two specific heuristics as side information tailored to language tasks: token frequencies and TF-IDF values of input tokens (or features).  These statistics are always known as open knowledge, thus can be used as an approximate how important each feature is. We compare \name with DP-SGD and DP-Adam described before.

\textbf{$A^t$ Based on Token Frequencies.} One easily obtainable side information is token frequencies, and we can set $A^t~(t \in [T])$ to be proportional to that accordingly. Note that our implicit assumption here is that rare features are more informative than frequent ones. We investigate the logistic regression model on two datasets in Figure 4 below. Despite the simplicity, this simple method works well on StackOverflow and IMDB under a tight privacy budget, outperforming DP-SGD and DP-Adam significantly. Especially for StackOverflow, the test accuracy is the same as that of \name {with} a small set of public data (Figure~\ref{fig:with-public-private}).

\begin{figure}[h!]
    \centering
    \includegraphics[width=0.49\textwidth]{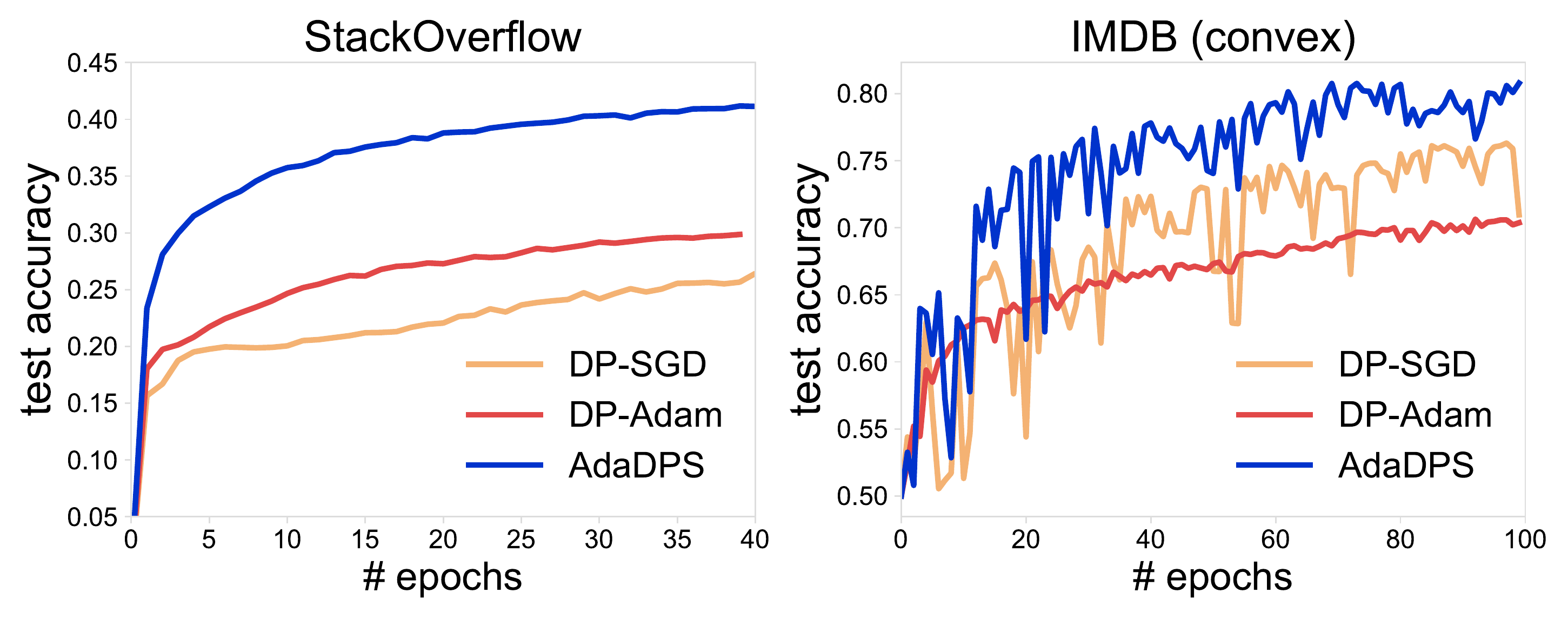}
    \caption{\name uses token frequencies as the side information, demonstrating superior performance than the baselines. Methods in each subfigure reach $(0.84, 4.2\times 10^{-6})$- and $(2.8, 4\times 10^{-5})$-DP. }
    \label{fig:without-public-private}
\end{figure}

\textbf{$A^t$ Based on TF-IDF Values.} Another common criterion to measure the relevance of tokens to each data point in a collection of training data is TF-IDF values. With the presence of such information available in a data processing pipeline, we explore the effects of $A^t$ being inversely proportional to TF-IDF scores for linear models.  The results are reported in Table~\ref{table: imdb_tf-idf}. The `ideal' method refers to \name with RMSProp updates and the second moment estimated on clean gradients from private training data, which serves a performance upper bound. As expected, across all methods, TF-IDF features result in higher accuracies than BoW features. \name outperforms the baselines by $\sim$10\% absolute accuracy, only slightly underperforming the `ideal' oracle.

\begin{table}[h!]
    \centering
    \scalebox{0.95}{
    \begin{tabular}{@{} l c c c c @{}}
    	\toprule[\heavyrulewidth]
    {\multirow{2}{*}{\textbf{Features}}} & \multicolumn{4}{c}{\textbf{Methods}} \\
         & DP-SGD  & DP-Adam & \name{} &  \textit{ideal}  \\
         \hline
          BoW & 0.62 {\small (.02)} & 0.68 {\small (.01)} &  \textbf{0.75} {\small (.01)} & \textit{0.82 {\small (.01)}} \\ 
          TF-IDF & 0.68 {\small (.01)} & 0.65 {\small (.01)}  &  \textbf{0.80} {\small (.00)} & \textit{0.83 {\small (.00)}} \\
          \bottomrule
    \end{tabular}}
    \caption{\small We preprocess IMDB into two versions with either BoW or TF-IDF features, and report average test accuracy along with standard deviation across three runs. \name with $A^t$ being inversely proportional to features' TF-IDF values outperforms the baselines of DP-SGD and DP-Adam by a large margin. \name also performs relatively closely to the `ideal' upper bound.}
    \label{table: imdb_tf-idf}
\end{table}

\textbf{Remark (Side Information in Non-Private Training).} The idea of using side information (with or without public data) can also improve the performance of vanilla SGD in non-private training, yielding similar accuracies as that of adaptive methods. We report additional results along this line in Table~\ref{table: dp-side_non_private} in the appendix.

\subsection{Federated Learning} \label{sec:exp:fl}
In this section, we discuss \name adapted to federated learning (FL) (learning statistical models over heterogeneous networks while keeping all data local) to satisfy user-level, global differential privacy (assuming a trusted central server). The canonical objective for FL is to fit a single model $w \in \mathbb{R}^d$ to data across a network of $n$ devices:
\medmuskip=0.5mu
\thinmuskip=0.5mu
\thickmuskip=0.5mu
\begin{align}\label{obj:fl}
    \min_{w\in \mathbb{R}^d} F(w) = \sum_{i=1}^n p_i f_i(w),
\end{align}
where $f_i(w)$ is the empirical local loss on each device $i$, and $p_i$ is some pre-defined weight for device $i$ such that $\sum_{i=1}^n p_i = 1$, which can be  $\frac{1}{n}$ or proportional to the number of local samples. In this work, we simply set $p_i=\frac{1}{n}$, $i\in [n]$. For this privacy-sensitive application, we assume there is no public data available. 

Due to the practical constraints of federated settings (e.g., unreliable network conditions, device heterogeneity, etc), federated optimization algorithms typically randomly samples a small subset of devices at each communication round, and allows each device to run optimization methods locally (e.g., local mini-batch SGD) before sending the updates back to the server~\cite{mcmahan2017communication}. Adapting \name to federated learning is not straightforward, raising questions of applying preconditioning at the server side, the device side, or both~\cite{wang2021field}.  We empirically find that on the considered dataset, preconditioning the mini-batch gradients \textit{locally at each iteration} demonstrates superior performance than preconditioning the entire model updates at the server side. The exact algorithm is summarized in Algorithm~\ref{alg:dp-side-fl} in the appendix.

We investigate the same StackOverflow dataset described in Section~\ref{sec:exp:centralized}, but follow its original, natural partition (by Stack Overflow users) for federated settings, one device per user. There are 400 devices in total for the subsampled version we use. We select 20 devices at each communication round and use a noise multiplier $\sigma=0.3$. While we arrive at a large $\varepsilon$ value for user-level DP, the final model could still be useful for defending against membership inference attacks in practice~\cite{kairouz2021practical}.

\begin{figure}[h!]
    \centering
    \includegraphics[width=0.48\textwidth]{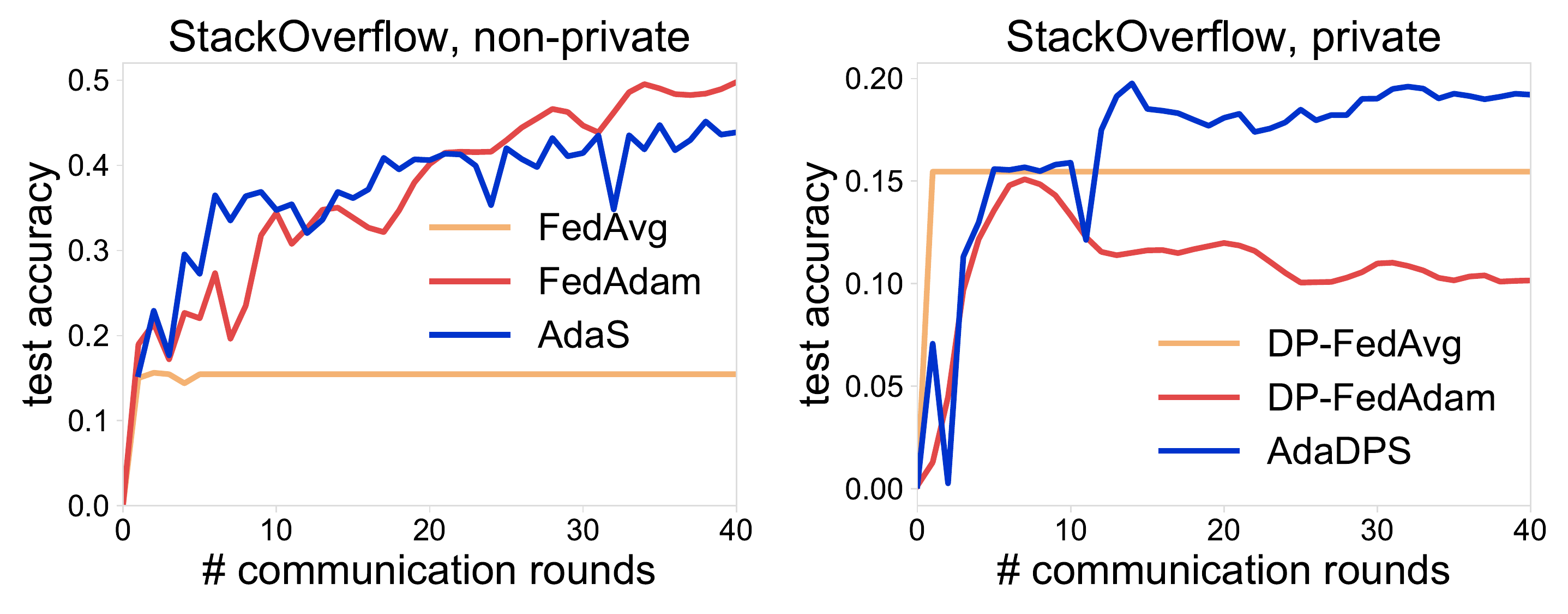}
    \caption{\small Test accuracy of StackOverflow in non-private and private settings under $(34,0.0025)$ user-level DP (Definition~\ref{def:user_dp}). \name extended to federated learning (Algorithm~\ref{alg:dp-side-fl} in the appendix) improves over baselines of DP-FedAvg~\cite{mcmahan2017learning} and DP-FedAdam by 5\% in terms of test accuracy.}
    \label{fig:so_fed}
\end{figure}

In Figure~\ref{fig:so_fed}, we plot test accuracy versus the number of communication rounds. \name has $\sim$5\% higher test accuracy than other two methods. 
We note that in federated learning applications involving massive and unreliable networks, it is not always realistic to allow for uniform device sampling. Incorporating recent advances in DP without sampling~\citep[e.g.,][]{kairouz2021practical} to address this is left for future work.

\section{Conclusion} \label{sec:conclusion}
In this work, we explored a simple and effective framework, \name, to realize the benefits of adaptive optimizers in differentially private learning via side information. Such information is used to precondition gradients before privatizing them. We analyzed the benefits of \name in terms of reducing the effective noise to reach similar privacy bounds, and empirically validated its superior performance across various tasks in both centralized and federated settings.

\section*{Acknowledgements}
We thank Brendan McMahan and Abhradeep Thakurta for valuable discussions and feedback. 
The work of TL and VS was supported in part by the National Science Foundation Grant IIS1838017, a Google Faculty Award, a Facebook Faculty Award, the Private AI Collaborative Research Institute, and the CONIX Research Center. Any opinions, findings, and conclusions or recommendations expressed in this material are those of the author(s) and do not necessarily reflect the National Science Foundation or any other funding agency.

\bibliography{refs}
\bibliographystyle{icml2022}

\newpage
\appendix
\onecolumn

\section{Convergence}

\subsection{Proof for Theorem~\ref{thm: convex_convergence} (Convex cases)} \label{app:convex_proof}
Based on our assumptions, the updating rule becomes
\begin{align}
    g^t &\gets \frac{1}{b}\sum_{i\in B} \nabla f(x^i; w^t) \\
    \hat{g}^t &\gets \frac{1}{b} \sum_{j \in B'} \nabla f(x^j; w^t),~x^j \in x_{\text{pub}} \\
    v^t &\gets \beta^t v^{t-1} + (1-\beta^t) (\hat{g}^t)^2 \\
    A^t &\gets\sqrt{v^t}+\epsilon^t \\
    w^{t+1} &\gets w^t - \alpha^t \left(\frac{g^t}{A^t}+N\right), ~N\sim \frac{1}{b}\mathcal{N} ({0}, \sigma^2 C^2)
\end{align}

We extend the proof in~\citet{mukkamala2017variants} to stochastic, private cases with preconditioner estimated on public data. Based on the updating rule, we have
\begin{align}
    &\quad \left\|w^{t+1}-w^*\right\|^2_{A^t} \\ &= \left\|w^{t}-\alpha^t (A^t)^{-1} g^t -\alpha^t N -w^*\right\|^2_{A^t} \\
    &= \left\|w^{t}-w^*\right\|_{A^t}^2 + \left\|\alpha^t (A^t)^{-1} g^t + \alpha^t N\right\|^2_{A^t} - 2 \left\langle w^t-w^*, \alpha^t g^t + \alpha^t A^t N \right\rangle \\
    &= \left\|w^{t}-w^*\right\|_{A^t}^2 - 2\alpha^t \left\langle g^t, w^t-w^* \right\rangle + (\alpha^t)^2 \left\langle g^t, (A^t)^{-1} g^t \right\rangle - 2\alpha^t \langle w^t-w^*, A^t N \rangle + (\alpha^t)^2 \|N\|^2_{A^t} + 2(\alpha^t)^2 \langle g^t, N \rangle.
\end{align}
Rearranging terms gives
\begin{align}
    \left\langle g^t, w^t-w^* \right\rangle = \frac{\|w^t-w^*\|^2_{A^t} - \|w^{t+1}-w^*\|^2_{A^t}}{2 \alpha^t} + \frac{\alpha^t}{2} \left\langle g^t, (A^t)^{-1} g^t \right\rangle - \langle w^t -w^*, A^t N \rangle + \frac{\alpha^t}{2} \|N\|^2_{A^t} +  \alpha^t  \langle g^t, N \rangle.
\end{align}
Take expectation on both sides conditioned on $w^t$,
\begin{align}
    \left\langle \nabla F(w^t), w^t-w^* \right\rangle = \frac{\mathbb{E}_t\left[\|w^t-w^*\|^2_{A^t}\right] - \mathbb{E}_t[\|w^{t+1}-w^*\|^2_{A^t}]}{2 \alpha^t} + \frac{\alpha^t}{2} \mathbb{E}_t\left[\left\langle g^t, (A^t)^{-1} g^t \right\rangle \right] + \frac{\alpha^t}{2}\mathbb{E}_t\left[\|N\|^2_{A^t}\right],
\end{align}
where we have used the fact that $N$ is a zero-mean Gaussian variable independent of $g^t, w^t$.
Taking expectation on both sides and using the convexity of $F(\cdot)$:
\begin{align}
    \mathbb{E}[F(w^t)] - F(w^*) &\leq \frac{\mathbb{E}[\|w^t-w^*\|^2_{A^t}] - \mathbb{E}[\|w^{t+1}-w^*\|^2_{A^t}]}{2\alpha^t} + \frac{\alpha^t}{2} \mathbb{E}[\left\langle g^t, (A^t)^{-1} g^t \right\rangle] + \frac{\alpha^t}{2}\mathbb{E}\left[\|N\|^2_{A^t}\right].
\end{align}
Applying telescope sum, we have
\begin{align}
    &\sum_{t=1}^T \left(\mathbb{E}[F(w^t)] - F(w^*)\right) \\&\leq \frac{\|w^1-w^*\|^2_{A^1}}{2\alpha_1} + \sum_{t=2}^T \left(\frac{\mathbb{E}\left[\|w^t-w^*\|^2_{A^t}\right]}{2\alpha^t}-\frac{\mathbb{E}\left[\|w^t-w^*\|^2_{A^{t-1}}\right]}{2\alpha_{t-1}}\right) + \sum_{t=1}^T \frac{\alpha^t}{2} \mathbb{E}\left[\left\langle g^t, (A^t)^{-1} g^t \right\rangle\right] + \sum_{t=1}^T \frac{\alpha^t}{2} \mathbb{E}\left[\|N\|^2_{A^t}\right].
\end{align}
Let $\alpha^t = \frac{\alpha}{\sqrt{t}}$,
\begin{align}
     &\sum_{t=1}^T \left(\mathbb{E}[F(w^t)] - F(w^*)\right) \\ &\leq  \frac{\|w^1-w^*\|^2_{A^1}}{2\alpha} + \underbrace{\sum_{t=2}^T \frac{\mathbb{E}\left[\|w^t-w^*\|^2_{\sqrt{t}A^t - \sqrt{t-1}A^{t-1}}\right]}{2\alpha}}_{T_1} + \underbrace{\sum_{t=1}^T \frac{\alpha}{2\sqrt{t}} \mathbb{E}\left[\left\langle g^t, (A^t)^{-1} g^t \right\rangle\right]}_{T_2}+\sum_{t=1}^T \frac{\alpha}{2\sqrt{t}} \mathbb{E}\left[\|N\|^2_{A^t}\right].
\end{align}
Let $ 1-\frac{\gamma}{t} \geq \beta^t \geq 1-\frac{1}{t}$ for some $\gamma \in (0, 1]$ and $\sqrt{t} \epsilon^t \geq \sqrt{t-1}\epsilon^{t-1}$.
We first bound $T_1$.
Based on the relations between $v^t$ and $v^{t-1}$ and $\beta^t \geq 1-\frac{1}{t}$, we can prove for any $t, j$
\begin{align}
    \sqrt{t} (A_j^t) = \sqrt{t} \left(\sqrt{v^t_j}+\epsilon^t\right) = \sqrt{t} \left(\sqrt{\beta^t v_j^{t-1}+(1-\beta^t) (\hat{g}_j^t)^2}+\epsilon^t\right) \geq \sqrt{(t-1) v^{t-1}_j} + \sqrt{t-1} \epsilon^{t-1}.
\end{align}
So for any $j, t$,
\begin{align}
    \sqrt{t} A_j^t \geq \sqrt{t-1}A_{j}^{t-1}.
\end{align}
Hence,
\begin{align}
     \mathbb{E}\left[ \sum_{t=2}^T \|w^t-w^*\|^2_{\sqrt{t} A^t - \sqrt{t-1}A^{t-1}}\right] &= \mathbb{E}\left[ \sum_{t=2}^T \sum_{j=1}^d (w^t_j-w^*_j)^2 \left(\sqrt{t v^t_j} + \sqrt{t} \epsilon^t - \sqrt{(t-1) v^{t-1}_j}-\sqrt{t-1}\epsilon^{t-1}\right)\right] \\
     & =\mathbb{E}\left[ \sum_{j=1}^d \sum_{t=2}^T (w^t_j-w^*_j)^2 \left(\sqrt{t v^t_j} + \sqrt{t} \epsilon^t - \sqrt{(t-1) v^{t-1}_j}-\sqrt{t-1}\epsilon^{t-1}\right)\right] \\
     &\leq \mathbb{E}\left[ \sum_{j=1}^d D^2 \sum_{t=2}^T  \left(\sqrt{t v^t_j} + \sqrt{t} \epsilon^t - \sqrt{(t-1) v^{t-1}_j}-\sqrt{t-1}\epsilon^{t-1}\right)\right]  \\
     &= \mathbb{E}\left[\sum_{j=1}^d D^2 \left(\sqrt{T v^T_j} + \sqrt{T} \epsilon^T - \sqrt{v^1_j} - \epsilon^1 \right)\right].
\end{align}
We next bound $T_2$. We prove a variant of  Lemma 4.1 in \citet{mukkamala2017variants}. The major differences are in that we consider the stochastic case and estimating $v^t$ on public data. We prove the following inequality by induction:
\begin{align}
    \sum_{t=1}^T \mathbb{E}\left[\frac{(g^t_j)^2}{\sqrt{t v^t_j} + \sqrt{t} \epsilon^t}\right] \leq \frac{2(2-\gamma)}{a\gamma} \mathbb{E}\left[\sqrt{T v^T_j} + \sqrt{T} \epsilon^T \right], ~j \in [d].
\end{align}
For $t=1$,
\begin{align}
    \mathbb{E}\left[\frac{(g_j^1)^2}{\sqrt{v_j^t}+\epsilon^1}\right] = \mathbb{E}\left[\frac{(g_j^1)^2}{\sqrt{(1-\beta^1) (\hat{g}_j^1)^2}+\epsilon^1}\right] \leq \mathbb{E}\left[\frac{(\hat{g}_j^1)^2}{a\sqrt{(1-\beta^1) (\hat{g}_j^1)^2}+\epsilon^1}\right] \leq \mathbb{E}\left[\frac{\sqrt{(1-\beta^1)(\hat{g}_j^1)^2}+\epsilon^1}{a(1-\beta^1)}\right]. 
\end{align}
Suppose that the conclusion holds when $t=T-1$, i.e., for any $j \in [d]$,
\begin{align}
    \sum_{t=1}^{T-1} \mathbb{E}\left[\frac{(g_j^t)^2}{\sqrt{t v_j^t} + \sqrt{t}\epsilon^t}\right] \leq \frac{2(2-\gamma)}{a\gamma} \mathbb{E}\left[\sqrt{(T-1) v_j^{T-1}} + \sqrt{T-1} \epsilon^{T-1}\right].
\end{align}
In addition, combined with the fact that $v_j^T = \beta^T v_j^{T-1}+(1-\beta^T) (\hat{g}_j^T)^2$ and $\sqrt{T}\epsilon^T \geq \sqrt{T-1}\epsilon^{T-1}$, we have
\begin{align}
    \sqrt{(T-1) v_j^{T-1}} + \sqrt{T-1} \epsilon^{T-1} &\leq \sqrt{\frac{(T-1)v_j^T}{\beta^T}-\frac{(T-1)(1-\beta^T)(\hat{g}_j^T)^2}{\beta^T}} + \sqrt{T} \epsilon^T \\
    &\leq \sqrt{T v_j^T -\frac{(T-1)(1-\beta^T)(\hat{g}_j^T)^2}{\beta^T}} + \sqrt{T}\epsilon^T \\
    &\leq \sqrt{T v_j^T} - \frac{(T-1)(1-\beta^T)(\hat{g}_j^T)^2}{2\beta^T \left(\sqrt{T v_j^T} + \sqrt{T} \epsilon^T\right)} + \sqrt{T} \epsilon^T \\
    &\leq \sqrt{T v_j^T} + \sqrt{T} \epsilon^T - \frac{a (T-1)(1-\beta^T)(g_j^T)^2}{2\beta^T \left(\sqrt{T v_j^T} + \sqrt{T} \epsilon^T\right)}.
\end{align}
The third inequality comes from $\sqrt{a-b} \leq \sqrt{a}-\frac{b}{2\sqrt{a}}~(a\geq b)$ by letting $a$  be $T v_j^T$ and $b$ be $\frac{(T-1)(1-\beta^T)(\hat{g}_j^T)^2}{\beta^T}$.
Hence
\begin{align}
    \sum_{t=1}^T \mathbb{E}\left[\frac{(g_j^t)^2}{\sqrt{t v_j^t} + \sqrt{t}\epsilon^t}\right] &\leq \frac{2(2-\gamma)}{a\gamma} \mathbb{E}\left[\sqrt{T v_j^T} + \sqrt{T} \epsilon^T - \frac{a (T-1)(1-\beta^T)(g_j^T)^2}{2\beta^T \left(\sqrt{T v_j^T} + \sqrt{T} \epsilon^T\right)}\right] + \mathbb{E}\left[\frac{(g_j^T)^2}{\sqrt{T v_j^T}+\sqrt{T}\epsilon^T}\right] \\
    &\leq \frac{2(2-\gamma)}{a\gamma} \mathbb{E}\left[\sqrt{T v_j^T} + \sqrt{T} \epsilon^T \right] + \left(1-\frac{(2-\gamma)(T-1)(1-\beta^T)}{\gamma \beta^T}\right) \mathbb{E}\left[\frac{(g_j^T)^2}{\sqrt{T v_j^T}+\sqrt{T}\epsilon^T}\right] \\
    &\leq \frac{2(2-\gamma)}{a\gamma} \mathbb{E}\left[\sqrt{T v_j^T} + \sqrt{T} \epsilon^T \right].
\end{align}
We then bound $T_2$ as follows.
\begin{align}
    T_2 = \mathbb{E}\left[\sum_{t=1}^T \frac{\alpha^t}{2} \sum_{j=1}^d \frac{(g^t_j)^2}{\sqrt{v^t_j}+\epsilon^t}\right] = \frac{\alpha}{2} \mathbb{E}\left[\sum_{t=1}^T  \sum_{j=1}^d \frac{(g^t_j)^2}{\sqrt{ t v^t_j}+ \sqrt{t}\epsilon^t}\right] \leq \frac{\alpha}{2} \sum_{j=1}^d \frac{2(2-\gamma)}{a\gamma} \mathbb{E}\left[\sqrt{T v^T_j} + \sqrt{T} \epsilon^T \right].
\end{align}
Noting that
\begin{align}
    \frac{\left\|w^1-w^*\right\|^2_{A^1}}{2\alpha} \leq \frac{D^2}{2\alpha} \sum_{j=1}^d \left(\sqrt{v^1_j} + \epsilon^1\right),
\end{align}
combined with the bounds of $T_1, T_2$ yields
\begin{align}
    \sum_{t=1}^T \left(\mathbb{E}[F(w^t)-F(w^*)]\right) \leq \left(\frac{D^2}{2\alpha} + \frac{\alpha (2-\gamma)}{a\gamma}\right) \sum_{j=1}^d \mathbb{E}\left[\sqrt{T v^T_j} + \sqrt{T} \epsilon^T \right], 
\end{align}
which implies that
\begin{align}
    \min_{t \in [T]} \mathbb{E}[F(w^t)] - F(w^*) &\leq \left(\frac{D^2}{2\alpha} + \frac{\alpha (2-\gamma)}{a\gamma}\right) \frac{1}{T} \sum_{j=1}^d \mathbb{E}\left[\sqrt{T v^T_j} + \sqrt{T} \epsilon^T \right]+\frac{1}{T}\sum_{t=1}^T \frac{\alpha}{2\sqrt{t}} \mathbb{E}\left[\|N\|^2_{A^t}\right] \\
   & \leq \left(\frac{D^2}{2\alpha} + \frac{\alpha (2-\gamma)}{a\gamma}\right) \frac{1}{\sqrt{T}} \sum_{j=1}^d \mathbb{E}\left[\sqrt{v^T_j} +  \epsilon^T \right]+\frac{\alpha}{\sqrt{T}}\max_{t\in [T]} \mathbb{E}\left[\|N\|^2_{A^t}\right]. 
\end{align}
The first term is standard for the RMSprop optimizer, and the last term is due to noise added to ensure differential privacy. To guarantee overall $(\varepsilon, \delta)$-differential privacy by running $T$ total iterations, we set $\sigma^2 = O\left(\frac{b^2  T \log(1/\delta)}{n^2 \varepsilon^2}\right)$ and $T=O\left(\frac{n^2\varepsilon^2}{ \max_{t\in [T]} \mathbb{E}\left[\sum_{j=1}^d A^t_j\right] \log(1/\delta)}\right)$.  The convergence rate becomes
\begin{align}
    \min_{t\in [T]}\mathbb{E}[F(w^t)] - F(w^*) \leq O\left(\frac{\sqrt{\max_{t\in [T]} \mathbb{E}\left[\sum_{j=1}^d A^t_j\right] \log(1/\delta)}}{n\varepsilon}\right).
\end{align}

\subsubsection{Proof for Theorem~\ref{thm:fixed_A} (Fix $A^t$ before training)}

Denote the side information as a fixed ${A}$ at any iteration $t$. Similar as previous analysis, setting a decaying learning rate $\alpha^t = \frac{\alpha}{\sqrt{t}}$, we have
\begin{align}
     &\sum_{t=1}^T \left(\mathbb{E}[F(w^t)] - F(w^*)\right) \\ &\leq  \frac{\|w^1-w^*\|^2_{{A}}}{2\alpha} + \underbrace{\sum_{t=2}^T \frac{\mathbb{E}\left[\|w^t-w^*\|^2_{\sqrt{t} {A} - \sqrt{t-1} {A}}\right]}{2\alpha}}_{T_1} + \underbrace{\sum_{t=1}^T \frac{\alpha}{2\sqrt{t}} \mathbb{E}\left[\left\langle g^t, ({A})^{-1} g^t \right\rangle\right]}_{T_2}+\sum_{t=1}^T \frac{\alpha}{2\sqrt{t}} \mathbb{E}\left[\|N\|^2_{{A}}\right].
\end{align}
To bound $T_1$, we have
\begin{align}
    \sum_{t=2}^T \mathbb{E}\left[\|w^t-w^*\|^2_{\sqrt{t}{A} - \sqrt{t-1}{A}}\right] = \mathbb{E}\left[\sum_{t=2}^T \sum_{j=1}^d (w_j^t-w_j^*)^2 \left((\sqrt{t}-\sqrt{t-1})({A}_j)\right)\right] \leq O\left(\sqrt{T} \sum_{j=1}^d {A}_j \right) - \left\|w^1-w^*\right\|_A^2.
\end{align}
We consider $T_2$ next. From the assumptions on the clipping bound,
\begin{align}
    R := \max_{j,t} \frac{\mathbb{E}\left[(g_j^t)^2\right]}{{A}_j^2} \leq C^2.
\end{align}
Then
\begin{align}
    \sum_{t=1}^T \sum_{j=1}^d \frac{\alpha}{2\sqrt{t} {A}_j} \mathbb{E}\left[(g^t_j)^2\right] \leq  \sum_{t=1}^T \sum_{j=1}^d \frac{\alpha}{2\sqrt{t}} R {A}_j \leq \sqrt{T} \alpha R \sum_{j=1}^d {A}_j.
\end{align}
Therefore, we obtain
\begin{align}
     \min_{t \in [T]} \mathbb{E}[F(w^t)] - F(w^*) \leq O\left(\frac{1}{\sqrt{T}} \sum_{j=1}^d {A}_j + \frac{\alpha R}{\sqrt{T}} \sum_{j=1}^d {A}_j + \frac{\alpha}{\sqrt{T}} \mathbb{E}\left[\|N\|^2_{{A}}\right]\right). \label{eq:2}
\end{align}


\subsection{Proof for Theorem~\ref{thm:convergence_non_convex} (Non-convex and smooth cases)} \label{app:non-convex_proof}

We use the same $\epsilon$ at each iteration.
Let $\nabla_j F(w)$ denote the $j$-th coordinate of $\nabla F(w)$ for any $w$. Based on the $L$-smoothness of $F(\cdot)$,
\begin{align}
    F(w^{t+1}) \leq F(w^t) &- \alpha^t \sum_{j=1}^d \left(\nabla_j F(w^t) \cdot \left(\frac{g^t_j}{\sqrt{v^t_j} + \epsilon} + N \right)\right) + \frac{(\alpha^t)^2 L}{2} \sum_{j=1}^d \left(\frac{(g^t_j)^2}{\left(\sqrt{v^t_j}+\epsilon\right)^2} + N^2 + \frac{g^t_j}{\sqrt{v^t_j} + \epsilon} \cdot 2N \right),
\end{align}
where $N \sim \frac{1}{b}\mathcal{N}(0, \sigma^2 C^2)$ and $\mathbb{E}[N^2]=\frac{\sigma^2 C^2}{b^2}$.
Taking expectation conditioned on $w^t$ on both sides gives
\begin{align}
    \mathbb{E}_t[F(w^{t+1})] \leq F(w^t) - \alpha^t \sum_{j=1}^d \nabla_j F(w^t) \mathbb{E}_t\left[\frac{g_j^t}{\sqrt{v_j^t}+\epsilon}\right] + \frac{(\alpha^t)^2 L}{2} \sum_{j=1}^d \mathbb{E}_t\left[\frac{(g_j^t)^2}{\left(\sqrt{v_j^t}+\epsilon\right)^2}\right] + \frac{(\alpha^t)^2 Ld}{2b^2}  \sigma^2 C^2.
\end{align}
The following proof is extended from that of Theorem 1 in~\citet{reddi2018adaptive}.
\begin{align}
    \mathbb{E}_t[F(w^{t+1})]  &\leq F(w^t) - \alpha^t \sum_{j=1}^d \nabla_j F(w^t) \mathbb{E}_t\left[\frac{g_j^t}{\sqrt{v_j^t}+\epsilon} - \frac{g_j^t}{\sqrt{\beta v_j^{t-1}}+\epsilon}  + \frac{g_j^t}{\sqrt{\beta v_j^{t-1}}+\epsilon} \right] \nonumber \\
    & \quad+\frac{(\alpha^t)^2 L}{2} \sum_{j=1}^d \mathbb{E}_t\left[\frac{(g_j^t)^2}{\left(\sqrt{v_j^t}+\epsilon\right)^2}\right] + \frac{(\alpha^t)^2 Ld}{2b^2}  \sigma^2 C^2 \\
    &\leq F(w^t) - \alpha^t \sum_{j=1}^d \frac{(\nabla_j F(w^t))^2}{\sqrt{\beta v_j^{t-1}}+\epsilon} + \alpha^t \sum_{j=1}^d \left|\nabla_j F(w^t)\right| \left|\mathbb{E}_t \left[\frac{g_j^t}{\sqrt{v_j^t}+\epsilon} - \frac{g_j^t}{\sqrt{\beta v_j^{t-1}}+\epsilon}\right]\right| \nonumber \\ 
    & \quad + \frac{(\alpha^t)^2 L}{2} \sum_{j=1}^d \mathbb{E}_t\left[\frac{(g_j^t)^2}{\left(\sqrt{v_j^t}+\epsilon\right)^2}\right] + \frac{(\alpha^t)^2 Ld}{2b^2}  \sigma^2 C^2.
\end{align}
Further,
\begin{align}
    \frac{g_j^t}{\sqrt{v_j^t}+\epsilon} - \frac{g_j^t}{\sqrt{\beta v_j^{t-1}}+\epsilon} &\leq \left|g_j^t\right| \left|\frac{1}{\sqrt{v_j^t}+\epsilon} - \frac{1}{\sqrt{\beta v_j^{t-1}}+\epsilon}\right| \\
    &= \frac{\left|g_j^t\right|}{\left(\sqrt{v_j^t}+\epsilon\right)\left(\sqrt{\beta v_j^{t-1}}+\epsilon \right)} \frac{(1-\beta) (\hat{g}_j^t)^2}{\left(\sqrt{v_j^t}+\sqrt{\beta v_j^{t-1}}\right)} \\
    &= \frac{\left|g_j^t\right|}{\left(\sqrt{v_j^t}+\epsilon\right)\left(\sqrt{\beta v_j^{t-1}}+\epsilon \right)} \frac{(1-\beta) (\hat{g}_j^t)^2}{\left(\sqrt{\beta v_j^{t-1} + (1-\beta) (\hat{g}_j^t)^2}+\sqrt{\beta v_j^{t-1}}\right)} \\
    &\leq \frac{1}{\left(\sqrt{v_j^t}+\epsilon\right)\left(\sqrt{\beta v_j^{t-1}}+\epsilon \right)} \frac{\sqrt{1-\beta}}{\sqrt{a}} (g_j^t)^2 \leq \frac{1}{\left(\sqrt{\beta v_j^{t-1}}+\epsilon\right)\epsilon} \frac{\sqrt{1-\beta}}{\sqrt{a}} (g_j^t)^2.
\end{align}
We have used the observation $\frac{(1-\beta) (\hat{g}_j^t)^2}{\left(\sqrt{\beta v_j^{t-1} + (1-\beta) (\hat{g}_j^t)^2}+\sqrt{\beta v_j^{t-1}}\right)} \leq \sqrt{1-\beta} \hat{g}_j^t$, and $\hat{g}_j^t \leq \frac{1}{\sqrt{a}} g_j^t$.

From $L$-smoothness of $F(\cdot)$ which implies that $\|\nabla F(u)-\nabla F(v) \| \leq L \|u-v\|$ for any $u, v \in \mathbb{R}^d$, and Assumption~\ref{assum:bounded_domain}, it is easy to see there exists a constant $B$ such that $|\nabla_j F(w)| \leq B$ for any $j \in [d]$.
\begin{align}
    &\mathbb{E}_t [F(w^{t+1})] \\ &\leq F(w^t) - \alpha^t \sum_{j=1}^d \frac{(\nabla_j F(w^t))^2}{\sqrt{\beta v_j^{t-1}} + \epsilon} + \frac{\alpha^t B\sqrt{1-\beta}}{\epsilon \sqrt{a}} \sum_{j=1}^d  \mathbb{E}_t\left[\frac{(g_j^t)^2}{\sqrt{\beta v_j^{t-1}} + \epsilon}\right] + \frac{(\alpha^t)^2 Ld}{2} \sum_{j=1}^d \mathbb{E}_t\left[\frac{(g_j^t)^2}{\left(\sqrt{v_j^t}+\epsilon\right)^2}\right] + \frac{(\alpha^t)^2 Ld}{2b^2}  \sigma^2 C^2 \\
    & \leq F(w^t) - \alpha^t \sum_{j=1}^d \frac{(\nabla_j F(w^t))^2}{\sqrt{\beta v_j^{t-1}} + \epsilon} + \frac{\alpha^t B\sqrt{1-\beta}}{\epsilon \sqrt{a}} \sum_{j=1}^d  \mathbb{E}_t\left[\frac{(g_j^t)^2}{\sqrt{\beta v_j^{t-1}} + \epsilon}\right] + \frac{(\alpha^t)^2 L}{2\epsilon} \sum_{j=1}^d \mathbb{E}_t \left[\frac{(g_j^t)^2}{\sqrt{\beta v_j^{t-1}} + \epsilon}\right] + \frac{(\alpha^t)^2 Ld}{2b^2}  \sigma^2 C^2,  \label{eq:4}
\end{align}
where the last inequality holds due to $\left(\sqrt{v_j^t}+\epsilon\right)^2 \geq \epsilon \left(\epsilon+\sqrt{v_j^t}\right) \geq \epsilon \left(\epsilon+\sqrt{\beta v_j^{t-1}}\right)$.
Lemma 1 in~\citet{reddi2018adaptive} proves that $\mathbb{E}_t\left[(g^{t}_j)^2\right] \leq \frac{\tau_j^2}{b} + (\nabla_j F(w^t))^2$ where $\tau_j^2$ is the variance bound of the $j$-th coordinate, i.e., $\mathbb{E}\left[\|g^t_j-\nabla_j F(w^t)\|^2\right] \leq \tau_j^2$. Plugging this inequality into Eq.~\eqref{eq:4}, combined with $\frac{L \alpha^t}{2\epsilon} \leq \frac{1}{4}$ and $\frac{B\sqrt{1-\beta}}{\sqrt{a}\epsilon} \leq \frac{1}{4}$, we obtain
\begin{align}
    \mathbb{E}_t[F(w^{t+1})] \leq F(w^t) - \frac{\alpha^t}{2 (\sqrt{\beta} B + \epsilon)} \|\nabla F(w^t)\|^2 + \left(\frac{\alpha^t B \sqrt{1-\beta}}{\sqrt{a} \epsilon^2}+\frac{L(\alpha^t)^2}{2\epsilon^2 \sqrt{\beta}}\right) \frac{\sum_{j\in [d]}\tau_j^2}{b} + \frac{(\alpha^t)^2 Ld}{2b^2} \sigma^2 C^2.
\end{align}
Taking expectation on both sides and applying the telescope sum yields
\begin{align}
    \frac{1}{T} \sum_{t \in [T]} \mathbb{E}[\|\nabla F(w^t)\|^2] \leq O\left(\frac{1}{T}\right) + O\left(\frac{ \|\tau^2\|_1}{b}\right) + O\left(\frac{\sigma^2}{b^2}\right)
\end{align}

\subsubsection{Proof for Theorem~\ref{thm:non-convex_fix_A} (Fix $A$ before training)}
Due to $L$-smoothness of $F(\cdot)$, we have
\begin{align}
     F(w^{t+1}) \leq F(w^t) &- \alpha^t \sum_{j=1}^d \left(\nabla_j F(w^t) \cdot \left(\frac{g^t_j}{A_j} + N \right)\right) + \frac{(\alpha^t)^2 L}{2} \sum_{j=1}^d \left(\frac{(g^t_j)^2}{A_j^2} + N^2 + \frac{g^t_j}{A_j} \cdot 2N\right),
\end{align}
where $N \sim \frac{1}{b}\mathcal{N}(0, \sigma^2 C^2)$. Taking expectation conditioned on $w^t$ on both sides gives
\begin{align}
    \mathbb{E}_t\left[F(w^{t+1})\right] &\leq F(w^t) - \alpha^t \sum_{j=1}^d \frac{(\nabla_j  F(w^t))^2}{{A}_j}  + \frac{(\alpha^t)^2 L}{2} \sum_{j=1}^d  \frac{1}{{A}_j^2} \mathbb{E}_t \left[(g_j^t)^2 \right] + \frac{(\alpha^t)^2 L d}{2b^2}  \sigma^2 C^2 \\
    &\leq F(w^t) - \alpha^t \sum_{j=1}^d \frac{(\nabla_j F(w^t))^2}{A_j} + \frac{(\alpha^t)^2 L}{2} \sum_{j=1}^d \frac{1}{A_j^2} \left(\frac{\tau_j^2}{b} + (\nabla_j F(w^t))^2\right) + \frac{(\alpha^t)^2 Ld}{2b^2} \sigma^2 C^2 \\
    &\leq F(w^t) - \alpha^t \sum_{j=1}^d \frac{(\nabla_j F(w^t))^2}{A_j} + \frac{(\alpha^t)^2 L}{2\epsilon} \sum_{j=1}^d \frac{1}{A_j} \left(\frac{\tau_j^2}{b} + (\nabla_j F(w^t))^2\right) + \frac{(\alpha^t)^2 Ld}{2b^2} \sigma^2 C^2 \\
    &\leq F(w^t) - \frac{\alpha^t}{2} \left\|\nabla F(w^t)\right\|^2_{A^{-1}} + \frac{(\alpha^t)^2 L}{2\epsilon} \sum_{j=1}^d \frac{\tau_j^2}{A_j b} + \frac{(\alpha^t)^2 Ld}{2b^2} \sigma^2 C^2.
\end{align}
The last inequality is due to $\alpha^t \leq \frac{\epsilon}{L}$.
Taking expectation on both sides yields
\begin{align}
    \mathbb{E}\left[F(w^{t+1})\right] \leq \mathbb{E}\left[F(w^t)\right] -\frac{\alpha^t}{2} \mathbb{E}[\|\nabla F(w^t)\|_{A^{-1}}^2] +  \frac{(\alpha^t)^2 L}{2\epsilon} \sum_{j=1}^d \frac{\tau_j^2}{A_j b} + \frac{(\alpha^t)^2 Ld}{2b^2} \sigma^2 C^2.
\end{align}
Similarly, by rearranging terms and applying telescope sum, we obtain
\begin{align}
    \frac{1}{T} \sum_{t\in[T]} \mathbb{E}\left[ \|\nabla F(w^t)\|_{A^{-1}}^2\right] \leq O\left(\frac{1}{T}\right) + O\left(\frac{\tau^2}{A}\cdot \frac{1}{b} \right) + O\left(\frac{\sigma^2}{b^2}\right).
\end{align}

\section{Experimental Details} \label{app:exp_details}

\textbf{Pseudo Code of some Algorithms.} For completeness, we present the full baseline DP-Adam algorithm in Algorithm~\ref{alg:dp-adam} and \name extended to federated learning in Algorithm~\ref{alg:dp-side-fl}.

\begin{algorithm}[h!]
\SetAlgoLined
\DontPrintSemicolon
\SetNoFillComment
\setlength{\abovedisplayskip}{3pt}
\setlength{\belowdisplayskip}{3pt}
\setlength{\abovedisplayshortskip}{3pt}
\setlength{\belowdisplayshortskip}{3pt}
\KwIn{$T$, batch size $b$, noise multiplier $\sigma$, clipping threshold $C$, initial model $w^1 \in \mathbb{R}^d$, $v^0=\mathbf{0}$, $m^0=\mathbf{0}$, small constant vector $\epsilon^t \in \mathbb{R}^d$, learning rate $\alpha^t$, moving average parameters $\beta_1$, $\beta_2$}
\caption{DP-Adam~\cite{zhou2020private}}
\label{alg:dp-adam}
    \For{$t=1, \cdots, T-1$}{
        Uniformly randomly sample a mini-batch $B$ with size $b$ from private training data \; 
        Get individual gradients for sample $i \in B$:
        \begin{align}
            g^{i,t} \gets \nabla f(x^i; w^t) \nonumber
        \end{align} \;
        Private gradients using Gaussian mechanism:
        \begin{align*}
            \Tilde{g}^t \gets \frac{1}{b} \sum_{i \in B} \text{clip} \left(g^{i,t}, C\right) + \frac{1}{b} \mathcal{N}({0}, \sigma^2 C^2)
        \end{align*} \\
        Update first and second moment estimates as
        \begin{align*}
            m^t &\gets \beta_1 m^{t-1} + (1-\beta_1) \Tilde{g}^t \\
            v^t &\gets \beta_2 v^{t-1} + (1-\beta_2) (\Tilde{g}^t)^2
        \end{align*} \\
        Update the model parameter $w$ as
        \begin{align*}
            w^{t+1} \gets w^t - \alpha^t \frac{m^t / (1-\beta_1^t)}{\sqrt{v^t / (1-\beta_2^t)}+\epsilon^t},
        \end{align*}
        where $\beta_1^t, \beta_2^t$ denotes $\beta_1, \beta_2$ to the power of $t$ (with slight abuse of notations)
        
        }
    \Return{$w^T$} \;
\end{algorithm}

\begin{algorithm}[h!]
\SetAlgoLined
\DontPrintSemicolon
\SetNoFillComment
\setlength{\abovedisplayskip}{3pt}
\setlength{\belowdisplayskip}{3pt}
\setlength{\abovedisplayshortskip}{3pt}
\setlength{\belowdisplayshortskip}{3pt}
\KwIn{$T$ communication rounds, $b$ selected devices each round, noise multiplier $\sigma$, clipping threshold $C$, initial model $w^1 \in \mathbb{R}^d$, non-sensitive side information $A$, number of local iterations $s$, local learning rate $\eta^t$}
\caption{\name applied to federated learning}
\label{alg:dp-side-fl}
    \For{$t=1, \cdots, T-1$}{
        Server uniformly selects a subset $B$ of $b$ devices and sends the current global model $w^t$ to them \;
        Each device $i \in B$ sets the local model to be the current global model: 
        \begin{align*}
            w^{i,0} \gets w^t 
        \end{align*} \\
        {Each device $i \in B$ runs adaptive mini-batch SGD locally with side information $A$ to obtain model updates: \\
        \For{$j=0, \cdots, s$}{
            \begin{align*}
            w^{i, j+1} \gets w^{i,j} - \eta^t \frac{\nabla f(w^{i,j})}{A}
            \end{align*}
            }
        And then privatize model updates:
            \begin{align*}
                \Delta^{i,t} &\gets w^{i, s+1}-w^{i,0} \\
                \Tilde{\Delta}^{i,t} &\gets \text{clip}(\Delta^{i,t}, C) + \mathcal{N}({0}, \sigma^2 C^2) 
            \end{align*}} \\
        Each device $ i \in B$ sends $\Tilde{\Delta}^{i,t}$ to the server \;
        Server updates the global model as:
        \begin{align*}
            w^{t+1} \gets w^t + \frac{1}{b} \sum_{i \in B} \Tilde{\Delta}^{i,t} 
        \end{align*}
        }
    \Return{$w^T$} \;
\end{algorithm}

\newpage
\textbf{Datasets and Models.} We consider a diverse set of datasets and tasks. 
\begin{itemize}[leftmargin=*]
    \item \textbf{StackOverflow}~\cite{stackoverflow} consists of posts on the Stack Overflow website, where the task is tag prediction (500-class classification). We randomly subsample 246,092 samples from the entire set. There are 10,000 input features in StackOverflow, resulting in more than 5 million learnable parameters in a logistic regression model. 
    \item \textbf{IMDB}~\cite{maas2011learning} is widely used for for binary sentiment classification of movie reviews, consisting of 25,000 training and 25,000 testing samples. We study two models on IMDB: logistic regression and neural networks (with LSTM) with 20,002 and 706,690 parameters, respectively. For logistic regression, we set the vocabulary size to 10,000 and consider two sets of commonly-used features separately: bag-of-words (BoW) and TF-IDF values. 
    \item  \textbf{MNIST}~\cite{lecun1998gradient} images with a deep autoencoder model (for image reconstruction) which has the same architecture as that in previous works~\cite{reddi2018adaptive} (containing more than 2 million parameters). The loss is reconstruction error measured as the mean squared distance in the pixel space. We scale each input feature to the range of $[0,1]$.
\end{itemize}

\paragraph{Hyperparameter Tuning.} We detail our hyperparameter tuning protocol and the hyperparameter values here. Our code is publicly available at~\href{https://github.com/litian96/AdaDPS}{\texttt{github.com/litian96/AdaDPS}}.

\begin{itemize}[leftmargin=*]
    \item For non-private training experiments, we fix the mini-batch size to 64, and tune fixed learning rates by performing a grid search over $\{0.0005, 0.001, 0.005, 0.01, 0.05, 0.1, 0.2, 0.5, 1, 2\}$ separately for all methods on validation data. We do not use momentum for AdaS (i.e., applying the idea of preconditioning of \name without privatization) for all centralized training experiments. 
    \item For differentially private training, the $\delta$ values in the privacy budget are always inverse of the number of training samples. We fix the noise multiplier $\sigma$ for each dataset, tune the clipping threshold, and compute the final $\varepsilon$ values. Specifically, the $\sigma$ values are 1, 1, and 0.95 for IMDB (convex), IMDB (LSTM), and StackOverflow; 1 and 0.75 for MNIST (autoencoder). The clipping threshold $C$ (in Algorithm~\ref{alg:dp-side}) is tuned from $\{0.01, 0.02, 0.05, 0.1, 0.2, 0.5, 1, 2, 3\}$, jointly with tuning the (fixed) learning rates. The number of micro-batches is 16 for all related experiments, and the mini-batch size is 64 (i.e., we privatize each gradient averaged over 4 individual ones to speed up computation).
    \item For federated learning experiments, we fix server-side learning rate to be 1 (i.e., simply applying the unscaled average of noisy model updates in Line 9 of Algorithm~\ref{alg:dp-side-fl}), and apply server-side momentum~\cite{reddi2020adaptive} with a moving average parameter $0.9$ for all methods in the left sub-figure in Figure~\ref{fig:so_fed}. The number of local epochs is set to 1 for all runs, and the local mini-batch size is 100.
\end{itemize}

The tuned hyperparameter values (clipping threshold $C$, learning rate) for private training are summarized in Table~\ref{table: hyperparameters} below. 

\begin{table}[h!]
    \centering
    \scalebox{0.92}{
    \begin{tabular}{@{} l  c  c c c  c @{}}
    \toprule[\heavyrulewidth]
       \textbf{Datasets} &  \textbf{DP-SGD} & \textbf{DP-Adam} & \textbf{DP-RMSProp} & \textbf{\name (RMSProp)} & \textbf{\name (Adam)}  \\
       \hline
        IMDB (convex) &  (0.1, 1) & (0.02, 0.1) & (0.05, 0.1) & (2, 0.5) & (2, 1) \\
        IMDB (LSTM) &  (0.1, 0.1) & (0.1, 0.001) & (0.1, 0.001) & (0.2, 0.1) & (0.2, 0.05) \\
        StackOverflow (linear) & (0.1, 1) & (0.1, 0.01) & (0.2, 0.01)  & (1, 0.5) & (1, 0.5) \\
        MNIST (autoencoder) & (0.05, 0.5) & (0.01, 0.001) & (0.01, 0.001) & (3, 0.005) & (3, 0.005) \\
        MNIST (classification) & (0.5, 0.01) & (0.5, 0.001) & (0.5, 0.001) & (2, 0.005) & (2, 0.005) \\
    \bottomrule
    \end{tabular}}
    \caption{\small Major hyperparameter values (learning rate and clipping threshold $C$) used in private experiments for all datasets. `IMDB (convex)' is IMDB (BoW features) on a logistic regression model. StackOverflow results are for centralized training. The noise multiplier $\sigma$ values in these four tasks are 1, 1, 0.95, 1, respectively, resulting in $\varepsilon$ values being 1.5, 2.8, 0.84, and 1.6.}
    \label{table: hyperparameters}
\end{table}

\section{Additional Results} \label{app:exp_additional}

\subsection{Additional Baselines}

\paragraph{Other Private Adaptive Optimization Baselines.} In the main text, we mainly compare \name with DP-Adam (summarized in Algorithm~\ref{alg:dp-adam}). There are other possible baselines similar as DP-Adam, by replacing Adam with other adaptive methods, resulting in DP-AdaGrad and DP-RMSProp. This line of differentially private optimizers has similar empirical performance as DP-Adam, as shown in the results in Table~\ref{table: all_methods} below (using DP-RMSProp as an example).

\begin{table}[h!]
    \centering
    \scalebox{0.84}{
    \begin{tabular}{@{} l l c  c c c  c @{}}
    \toprule[\heavyrulewidth]
       \textbf{Datasets} & \textbf{Metrics} & \textbf{DP-SGD} & \textbf{DP-Adam} & \textbf{DP-RMSProp} & \textbf{\name (RMSProp)} & \textbf{\name (Adam)}  \\
       \hline
        IMDB (convex) & accuracy & 0.63 & 0.69 & 0.67 & 0.80 & 0.80 \\
        IMDB (LSTM) & accuracy & 0.70 & 0.69 & 0.69 & 0.73 & 0.73 \\
        StackOverflow (linear) & accuracy  & 0.28 & 0.30 & 0.31  & 0.40 & 0.40 \\
        MNIST (autoencoder) & loss {\small ($\times$100)} & 5.013 & 3.697 & 3.636 & 3.566 & 3.443 \\
        MNIST (classification) & accuracy & 0.9273 & 0.9333 & 0.9314 & 0.9377 &  0.9541 \\
    \bottomrule
    \end{tabular}}
    \caption{\small Full comparisons between \name and private adaptive optimization methods. The evaluation metrics are reported on test data. `IMDB (convex)' is IMDB (BoW features) on a logistic regression model. For $(\varepsilon, \delta)$-differential privacy, the $\varepsilon$ values of experiments in the five rows are 1.5, 2.8, 0.84, 1.6, and 1.25, respectively, and the $\delta$ values are the inverse of the number of training samples (as mentioned in the main text).}
    \label{table: all_methods}
\end{table}

\paragraph{Using Public Data for Pretraining.} Another possible way of leveraging public data to improve privacy/utility tradeoffs is to pretrain on them. However, this would give only limited performance improvement if the amount of public data is very small. In the main text, when needed, \name randomly samples 1\% training data as public data. Under this setup, we empirically compare \name with the pre-training baseline (denoted as DP-SGD w/ warm start). From Table~\ref{table: compare_public_pretrain}, we see that \name outperforms it by a large margin.

\begin{table}[h!]
    \centering
    \scalebox{0.99}{
    \begin{tabular}{@{} l  l c c c @{}}
    \toprule[\heavyrulewidth]
    {\multirow{2}{*}{\textbf{Datasets}}} 
    & \multicolumn{1}{l}{\multirow{2}{*}{
       \textbf{Metrics}}} & \multicolumn{1}{l}{\multirow{2}{*}{
       \textbf{DP-SGD}}} & \textbf{DP-SGD} & \textbf{\name}    \\
              & & & w/ warm start & w/ public  \\
       \hline
        IMDB (convex)  & accuracy &  0.63    &  0.73   & 0.80  \\
        StackOverflow  & accuracy &  0.28    &  0.33   & 0.40  \\
        MNIST (autoencoder)          & loss     &  0.050   &  0.049  & 0.036 \\
    \bottomrule
    \end{tabular}}
    \caption{\small  Compare \name with an additional baseline of DP-SGD pre-trained on public data on three datasets. For `DP-SGD w/ warm start', we first train on public data for 10 epochs via adaptive methods (RMSProp), and then run DP-SGD on private data starting from that initialization.  
    }
    \label{table: compare_public_pretrain}
\end{table}

\paragraph{DP-Adam with Public Data.} In the main text (Section~\ref{sec:exp:centralized:with_public}), we discuss the DP-R-Pub. baseline based on the RMSProp method with the preconditioner estimated on public data. Similarly, one can also apply such clean preconditioners in DP-Adam updates, resulting in another baseline which we call DP-Adam-Pub. The main differences between DP-Adam-Pub and DP-R-Pub are that DP-Adam-Pub additionally considers momentum. Formally, the updates of $w^t$ is as follows:
\begin{align*}
   \Tilde{g}^t &\gets \frac{1}{b} \sum_{i \in B} \text{clip}\left(g^{i,t}, C\right)+\frac{1}{b}\mathcal{N}(0, \sigma^2 C^2), \quad \hat{g}^t \gets \mathbb{E}_x\left[\nabla f(x; w^t)\right] \text{ for } x \in x_{\text{pub}}, \\ 
   m^t &\gets \beta_2 m^t + (1-\beta_2) \left(\beta_1 \Tilde{g}^t + (1-\beta_1) \hat{g}^t\right), \quad v^t \gets \beta_3 v^t + (1-\beta_3) (\hat{g}^t)^2, 
   \\ w^{t+1} &\gets w^t - \alpha^t \frac{ m^t / (1-(\beta_2)^t) } {\sqrt{v^t / (1-(\beta_3)^t)}+\epsilon},
\end{align*}
where $\beta_1, \beta_2, \beta_3, \epsilon$ are small constants.

\begin{table}[h!]
    \centering
    \scalebox{0.99}{
    \begin{tabular}{@{} l  l c c c @{}}
    \toprule[\heavyrulewidth]
    {\multirow{2}{*}{\textbf{Datasets}}} 
    & \multicolumn{1}{l}{\multirow{2}{*}{
       \textbf{Metrics}}} & \multicolumn{1}{l}{\multirow{2}{*}{
       \textbf{DP-SGD}}} & \multicolumn{1}{l}{\multirow{2}{*}{
       \textbf{DP-Adam-Pub}}} & \textbf{\name}    \\
              & & &  & w/ public  \\
       \hline
        IMDB (convex)  & accuracy &  0.63    &  0.74   & 0.80  \\
        StackOverflow  & accuracy &  0.28    &  0.31   & 0.40  \\
        MNIST (autoencoder) & loss     &  0.050   &  0.064  & 0.036 \\
    \bottomrule
    \end{tabular}}
    \caption{\small  Results of comparing \name with to DP-Adam-Pub (i.e., DP-Adam using clean preconditioners estimated on public data).
    }
    \label{table: compare_dp-adam-pub}
\end{table}

\subsection{Side Information in Non-Private Training}
In the main text, we mainly focus on private optimization. It is expected that side information (even without the assist of public data) would also be beneficial in non-private settings, which could serve as a simple alternative to adaptive methods. We report results in Table~\ref{table: dp-side_non_private}.

\begin{table}[h!]
    \centering
    \begin{tabular}{@{} l l c  c c c @{}}
    \toprule[\heavyrulewidth]
       \textbf{Datasets}  & \textbf{Metrics} & \textbf{SGD} & \textbf{Adam} & \textbf{\texttt{AdaS}} (w/ public) & \textbf{\texttt{AdaS}} (w/o public)   \\
       \hline
        IMDB (convex) & accuracy  & 0.66 & 0.88  &  0.88 & 0.88 \\
        IMDB (LSTM) & accuracy & 0.88 & 0.88 & 0.88 & 0.88 \\
        StackOverflow (linear) & accuracy & 0.38 & 0.64 & 0.64 & 0.64 \\
        MNIST (autoencoder) & {\small loss ($\times$100)} & 5.013 & 1.151 & 1.805 & --- \\
    \bottomrule
    \end{tabular}
    \caption{\small Performance of each method in \textit{non-private} training. We see that \texttt{AdaS} can match the performance of Adam in non-private settings.}
    \label{table: dp-side_non_private}
\end{table}

\subsection{Effects of Public Data Size}

We further study the effects of public data size. Only a very small set of public data (even 0.04\% the size of private training data) can provide good preconditioner estimates.

\begin{table}[h!]
    \centering
    \begin{tabular}{@{} l  c c c c @{}}
    \toprule[\heavyrulewidth]
    {\multirow{2}{*}{\textbf{Datasets}}} 
     & \multicolumn{1}{l}{\multirow{2}{*}{
      {\textit{upper bound}}}} & \textbf{\name}  & \textbf{\name}  & \textbf{\name}   \\
              &  & 1\% public & 5$\times$ less & 25$\times$ less \\
       \hline
        IMDB (convex)   &   0.82  &   0.80      & 0.80  &  0.75 \\
        StackOverflow   &   0.41  &  0.40   & 0.40  &  0.39  \\
    \bottomrule
    \end{tabular}
    \caption{\small  Effects of public data sizes.}
    \label{table: public-data-size}
\end{table}

\end{document}